\documentclass[runningheads]{llncs}
\usepackage{graphicx}
\usepackage{amsmath,amssymb} 
\usepackage{color}
\usepackage{isomath}
\usepackage[T1]{fontenc}    
\usepackage{url}            
\usepackage{booktabs}       
\usepackage{amsfonts}       
\usepackage{nicefrac}       
\usepackage{microtype}      
\usepackage{xcolor}
\usepackage{epsfig}
\usepackage{multirow}
\usepackage{bbm}
\usepackage{bm}
\usepackage{textcomp}
\usepackage{footnote}
\usepackage{tablefootnote}
\usepackage[linesnumbered,ruled]{algorithm2e}
\usepackage{mathtools}
\usepackage{sidecap}
\usepackage{hyperref}

\makeatletter
\newcommand{\printfnsymbol}[1]{%
  \textsuperscript{\@fnsymbol{#1}}%
}
\makeatother
\begin{document}
\title{Coreset-Based Neural Network Compression}

\titlerunning{Coreset-Based Neural Network Compression}
%
\author{Abhimanyu Dubey\thanks{Equal Contribution.}\inst{1} \and
Moitreya Chatterjee\printfnsymbol{1}\inst{2} \and
Narendra Ahuja\inst{2}}
\authorrunning{A. Dubey$^*$, M. Chatterjee$^*$ and N. Ahuja}
%

\institute{Massachusetts Institute of Technology, Cambridge MA 02139, USA \\
\email{dubeya@mit.edu}\and
University of Illinois at Urbana-Champaign, Champaign IL 61820, USA\\
\email{metro.smiles@gmail.com, } \email{n-ahuja@illinois.edu}}
\maketitle              
\begin{abstract}
We propose a novel Convolutional Neural Network (CNN) compression algorithm based on coreset representations of filters. We exploit the redundancies extant in the space of CNN weights and neuronal activations (across samples) in order to obtain compression.  Our method requires no retraining, is easy to implement, and obtains state-of-the-art compression performance across a wide variety of CNN architectures. Coupled with quantization and Huffman coding, we create networks that provide AlexNet-like accuracy, with a memory footprint that is $832\times$ smaller than the original AlexNet, while also introducing significant reductions in inference time as well. Additionally these compressed networks when fine-tuned, successfully generalize to other domains as well.
\end{abstract}
\section{Introduction}
Convolutional neural networks, while immensely powerful, often are resource-intensive\cite{krizhevsky2012imagenet,simonyan2014very,szegedy2015going,he2015deep,he2016identity}. Popular CNN models such as AlexNet~\cite{krizhevsky2012imagenet} and VGG-16~\cite{simonyan2014very}, for instance, have 61 and 138 million parameters and consume in excess of 200MB and 500MB of memory space respectively. This characteristic of deep CNN architectures reduces their portability, and poses a severe bottleneck for implementation in resource constrained environments \cite{gokhale2014240}. Additionally, design choices for CNN architectures, such as network depth, filter sizes, and number of filters seem arbitrary and motivated purely by empirical performance at a particular task, permitting little room for interpretability. Moreover, the architecture design is not necessarily fully optimized for the network to be yielding a certain level of precision, making these models highly resource-inefficient.

Several prior approaches have thus sought to reduce the computational complexity of these models. Work aimed at designing efficient CNN architectures, such as Residual Networks (ResNets)~\cite{he2016deep} and DenseNets~\cite{huang2016densely} have shown promise at alleviating the challenge of model complexity. These CNNs provide higher performance on classification at only a fraction of the number of parameters of their more resource intensive counterparts. However, despite being more compact, redundancies remain in such networks, leaving room for further compression.

In this work, we propose a novel method that exploits inter-filter dependencies extant in the convolutional filter banks of CNNs to compress pre-trained computationally intensive neural networks. Additionally we leverage neuronal activation patterns across samples to prune out irrelevant filters. Our compression pipeline consists of finely pruning the filters of every layer of the CNN based on sample activation patterns, followed by the construction of efficient filter \textit{coreset} representations to exploit the inter-filter dependencies. Our method \textit{does not require retraining}, is applicable \textit{to both fully-connected and convolution layers}, and maintains classification performance similar to the uncompressed network. We display state-of-the-art compression rates on several popular CNN models, including multiple ResNets, which show increases from $9.2\times$ to $16.2\times$ in compression rate over prior state-of-the-art techniques. Coupled with \textit{Deep Compression}, we are additionally able to compress other popular CNN models such as VGGNet-16~\cite{simonyan2014very} and AlexNet~\cite{krizhevsky2012imagenet} by 238$\times$ and 55$\times$ respectively. Moreover, we demonstrate the presence of filter redundancies even in highly efficient models such as SqueezeNet~\cite{iandola2016squeezenet}, by reducing their parameters by 50\% with almost no loss in classification performance, giving us AlexNet-level precision but with \textit{832$\times$} smaller model size, compared to the original AlexNet model. Finally, we empirically validate the generalizability of these compressed CNNs to newer domains.

In the next section, we discuss relevant prior work in this area. In Section 3, we present the details of our algorithm. This is followed by Section 4, where a discussion on the empirical evaluation of our method vis-\`a-vis other competing compression techniques is presented. We finally conclude in Section 5, laying out some avenues for future research in this area.
\section{Related Work}
\textbf{Network Compression}: Compressing neural networks has been a topic of active research interest lately. Prior work in this area can be grouped into three distinct categories. The first category of methods direct their attention to the construction of parameter-efficient neural network architectures. For instance, Iandola~\textit{et al.}\cite{iandola2016squeezenet} propose SqueezeNets, a neural architecture class containing the parameter efficient, fully convolutional, `fire' modules. Other examples of such architectures include Residual Networks (ResNets)\cite{he2016deep}, and Densely Connected Neural Networks (DenseNets)\cite{huang2016densely}, which provide higher classification performance with models much smaller than the previous state-of-the-art, using `skip-connections' between layers of the network. More recent approaches have sought to adapt CNN architectures so as to make them robust to common transformations (e.g. rotation) within the data, by modifying the filter banks of a CNN~\cite{zhai2016doubly,dieleman2016exploiting} or by enforcing sparsity while training \cite{alvarez2017compression}. While these approaches seem to hold promise but they fail to fully exploit the inter-filter dependencies, allowing room for further compression of such networks. Meta-learning approaches attempt to decipher the optimum CNN architecture by searching over the space of a gigantic number of possible candidates. However, these techniques are prohibitively resource intensive, (needing well in excess of 400 GPUs to run), and often yield only a locally optimum architecture \cite{real2017large}.

A second broad category of compression methods attempt to prune the unimportant network parameters. Han~\textit{et al.}\cite{han2015deep} demonstrate an efficient pruning-retraining method, based on pruning weights by their $\ell^p$ norms. Srinivas and Babu~\cite{srinivas2015data} remove individual neurons instead of weights, with impressive results. The importance of ordering filters for the purpose of pruning has also been highlighted in Yu~\textit{et al.}, He~\textit{et al.}, and Molchanov~\textit{et al.}~\cite{yu2017nisp,he2017channel,molchanov2016pruning}. These approaches have been modified in the works of Polyak~\textit{et al.}\cite{polyak2015channel} and Luo~\textit{et al.}\cite{luo2017thinet}, that focus on removing weights grouped by characteristics of filters (such as norm of filter weights, etc.). Li~\textit{et al.}\cite{li2016pruning} extend the ideas of filter pruning by removing filters from a network following an `importance' criterion. However, the re-training step in these algorithms is time intensive.

Finally, the third theme of compression techniques is to employ weight-approximation and information-theoretic principles for the compression of neural network parameters. An early example of such work is the approach by Denton~\textit{et al.}\cite{denton2014exploiting} that uses low-rank approximations to compress fully-connected layers of neural networks. However, this technique doesn't apply to the convolution layers. Lebdev~\textit{et al.} fixes this problem and employs a low-rank decomposition approach to the full CNN to construct more efficient representations but their technique's principal bottleneck is re-training, which we avoid~\cite{lebedev2014speeding}. Rosenfield~\textit{et al.} consider an efficient utilization of CNN filters by representing them as a linear combination of a bases set~\cite{rosenfeld2017incremental}. However, our algorithm, different from this line of work, is additionally also capable of introducing structure, such as sparsity, in the approximated weights resulting from the decomposition, which further aids compression. Han~\textit{et al.}\cite{han2015deep} introduce \textit{Deep Compression}, that uses several steps such as weight-pruning, weight-sharing, and Huffman coding to reduce neural network size. However, their algorithm requires special hardware for inference in the compressed state, making it hard to deploy the compressed networks across platforms.

Our method aims at handling the shortcomings in each of these individual themes of CNN compression. Contrary to the first category of architecture search, our method is applicable to a wide variety of models, is less resource intensive, and does not require any retraining. While we do prune filters inspired by the work of Polyak~\textit{et al.}\cite{polyak2015channel} (following the second theme of compression), our criterion for filter pruning, however, is motivated by the accurate reconstruction of sample activations, instead of the magnitude of filter weights. Finally, our compression technique does not require special hardware for running inference unlike Han~\textit{et al.}\cite{han2016eie} and scales to both fully-connected and convolutional layers, unlike the low-rank (SVD) approach by Denton~\textit{et al.}\cite{denton2014exploiting}.

\textbf{Coresets for Point Selection:} Coresets have been widely studied in computational geometry. They were introduced first by Agarwal~\textit{et al.}\cite{agarwal2005geometric} for approximating a set of points with a smaller set, while preserving some desired criteria, on k-means and k-median problems. Badoiu~\textit{et al.}\cite{badoiu2002approximate} propose a coreset formulation to cluster points using a subset of the total set of points to generate the optimal solution. Har-Peled and Mazumdar~\cite{HarPeled2004} give an alternate solution for coresets that include points not in the original set. Feldman~\textit{et al.}\cite{feldman2007ptas} demonstrate that weak coreset representations can be generated with the number of points independent of the underlying data distribution. These formulations have recently been applied to several problems within computer vision and machine learning \cite{feldman2013turning,feldman2015dimensionality,dubey2015coreset}, and are primarily used to approximate a set of $n$ points in $d$ dimensions, originating from a domain $\mathbf{S}$, with a smaller set of $\tilde n << n$ points, while preserving some criterion such as similar pairwise distances. However, coresets have remained unexplored in the context of CNN compression, which constitutes a major novelty of our work.

\section{Method}
We begin with a fully-trained CNN and compress it without retraining, first by pruning out unimportant filters, followed by extraction of efficient coreset representation of these filters. Some of the major advantages of our method include: (i) Lack of retraining, therefore a major reduction in processing time, (ii) Capacity of our algorithm to significantly compress both convolutional and fully connected layers, and (iii) Ability of the compressed CNN to generalize to newer tasks.

\subsection{Background and Notation}
An $n$-layered neural network can be described as a union of the parameter tensors of every layer, $\mathcal W = \cup_{k=1}^n \ \tensorsym W_k$. The parameters $\tensorsym W_k$ of layer $k$ have the shape $N_k \times C_k \times h_k \times w_k$, where $N_k$ denotes the number of filters, $C_k$ denotes the number of input channels of the filter (since this is typically equal to the number of filters in the previous layer, $C_k = N_{k-1}$), and $h_k$ and $w_k$ denote the height and width of a filter. We can rewrite the parameter tensor $\tensorsym W_k$ as a 2D matrix $\vectorsym W_k$ of the shape $N_k \times (C_k h_k w_k)$. Next, we append the biases of the filters to $\vectorsym W_k$, to make it a matrix of dimensions $N_k \times (C_k h_k w_k + 1)$. It is well known that using this representation of the weights and biases of a layer, $\vectorsym W_k$, we can represent the output activation of any fully connected layer as a matrix product of $\matrixsym W_k$ with the incoming activation tensor $\tensorsym A_{k-1}$. This notion can be extended to convolution layers by re-casting the matrices in an appropriate Toeplitz form ~\cite{vasudevan2017parallel}.

The goal of compression is to obtain a compressed representation of the parameters for each layer $\hat{\mathcal W} = \cup_{k=1}^n \ \hat{\tensorsym W_k}$ such that it is smaller and computationally efficient, and preserves the final classification accuracy. Our approach is to construct compressed filter `coresets' $\hat{\vectorsym W_k} \in \mathbb R^{\hat{N}_k \times (C_k h_k w_k + 1)}$ of the parameters of each layer (where $\hat{N}_k < N_k$), such that the output activations (obtained after the Toeplitz matrix multiplication), are well approximated. Ensuring that the output activations remain largely the same at every layer, post compression, ensures that the final classification performance remains largely unchanged. Since the elements of these coresets are typically linear functions of the original parameters, we will additionally require a decompression matrix $\matrixsym D_k \in \mathbb R^{\hat{N}_k \times N_k}$ to obtain an approximation to the initial set of parameters, starting with the coreset representation.
%

Coresets are an effective technique of approximating a large set of points with a smaller set, which need not necessarily be a part the original set, while preserving some desirable property such as mean pairwise distances, diameter of the point set, etc. We seek to obtain a reduced matrix (coreset) $\hat{\matrixsym W_k}$ representation of the original filter weights $\matrixsym W_k$ of every layer, which we do via 3 different approaches, as described below.

\subsection{k-Means Coresets}
A first approach to constructing such a coreset would be to obtain a reduced representation of the parameter matrix that approximates the sum of distances in the space of neuronal activations of an arbitrary sample between each of the filters. Feldman~\textit{et al.}\cite{feldman2013turning} demonstrate that this problem is equivalent to finding a low-rank approximation of the filter matrix. This is representable as follows:
\begin{gather}
  \min_{\matrixsym U'_k, \matrixsym \Sigma'_k, \matrixsym V'_k} \lVert \matrixsym W_k - \matrixsym U'_k\matrixsym \Sigma'_k\matrixsym V'^{T}_k \rVert_F^2
\end{gather}

Their formulation for constructing a compact set of $\hat{N_k} << N$ points using the sum of distances criterion leads to the \textit{k-Means Coresets}, where the coreset representation is given by the solution to the above optimization problem:
\begin{equation}
 \hat{\matrixsym W_k} = \matrixsym U'_k \matrixsym \Sigma'_k, \text{ with decompression matrix: } \matrixsym D_k = \matrixsym V^{'T}_k
\end{equation}

Here, the matrices $\matrixsym U'_k, \matrixsym \Sigma'_k$ and $\matrixsym V^{'T}_k$ are the $\hat N_k$-truncated versions of the matrices $\matrixsym U_k, \matrixsym \Sigma_k$ and $\matrixsym V^T_k$, which satisfy the property:
\begin{gather}
  \matrixsym W_k = \matrixsym U_k \matrixsym \Sigma_k \matrixsym V^T_k \approx \hat{\matrixsym W_k} = \matrixsym U'_k \matrixsym \Sigma'_k \matrixsym V^{'T}_k
\end{gather}

$\matrixsym U_k, \matrixsym V_k$ are unitary matrices, while $\matrixsym \Sigma_k$ is a diagonal matrix. Such a decomposition can be obtained using Singular Value Decomposition (SVD), where the extent of truncation is specified as an input to the algorithm. The truncation determines the amount of compression we get.

Intuitively a significant truncation, while yielding greater compression, leads to a weaker approximation of the filter weights. This also results in a weaker approximation of the output activations, manifesting itself as a drop in classification accuracy. We seek the optimum compression, across all layers, such that the classification accuracy does not deviate by more than 0.5\%. 

SVD for compressing neural network weights has been investigated previously in~\cite{denton2014exploiting}, however, with two key differences - (i) the naive SVD approach has been applied only to fully-connected layers of neural networks, with limited success, whereas our coreset-based formulation scales to both convolution and fully connected layers, and (ii) our method for selecting the number of components to be retained $\hat{N_k}$ is data-dependent, based on training error obtained on random subsets of the training data, instead of an arbitrary initialization followed by retraining. However, since this decomposition does not explicitly encode any structure on the approximated weights, such as sparsity or considers the impact of activations, we build upon this formulation to create stronger coreset representations. This sets us apart from prior work, which employ simple low-rank decomposition for constructing efficient CNNs~\cite{lebedev2014speeding}.

\subsection{Structured Sparse Coresets}
If we consider the previous coreset decomposition, the optimization problem can be rewritten as (subject to constraints on each of the variables $\matrixsym U'_k, \matrixsym \Sigma'_k$ and $\matrixsym V'_k$):
\begin{gather}
  \min_{\matrixsym U'_k, \matrixsym \Sigma'_k, \matrixsym V'_k} \lVert \matrixsym W_k - \matrixsym U'_k\matrixsym \Sigma'_k\matrixsym V'^{T}_k \rVert_F^2
\end{gather}
To induce sparsity in the obtained decomposition, Jenatton~\textit{et al.}\cite{jenatton2010structured} introduce a technique known as \textit{Structured Sparse PCA}, which optimizes the following: 
\begin{equation}
\begin{split}
\min_{\matrixsym U'_k, \matrixsym \Sigma'_k, \matrixsym V'_k} \lVert \matrixsym W_k - \matrixsym U'_k\matrixsym \Sigma'_k\matrixsym V'^T_k \rVert_F^2 + \lambda \cdot \lVert V'_k \rVert_1, \\
\text{ subject to } \ \lVert (\matrixsym U'_k \cdot \Sigma'_k)_{m} \rVert_2 = 1 \ \forall \ m \in [1, \hat{N_k}]
\end{split}
\end{equation}
This problem can be solved by a cyclic optimization of two convex problems~\cite{jenatton2010structured}, and provides us with a decomposition that possesses structured sparsity. The motivation behind using such a formulation is to obtain a decomposition that is sparse in the number of components used, while minimizing reconstruction error. While techniques such as SPCA~\cite{jolliffe2003modified} or NMF~\cite{lee1999learning} also construct representations that are sparse in the projected space, this formulation returns a decomposition that makes the approximation in the original space sparse as well, hence, \textit{both} $\hat{\matrixsym W_k}$ and $\matrixsym D_k$ are sparse. Moreover, this formulation allows us to discard those filters for which the corresponding column vector in $\matrixsym D_k$ is a null vector, leading to further compression.

The hyper-parameters $\hat N_k$ and $\lambda$ are chosen jointly so as to obtain the maximum compression while restricting the deviation in classification performance to within 0.5\% of the uncompressed network, post the compression of all layers. We observe that this technique provides much more compression than k-Means Coreset, however, this does not take into account the relative importance of the filters during reconstruction, which leads us to our final coreset formulation.
\begin{figure*}[t]
  \centering
  \includegraphics[height = 3cm, width=0.8\textwidth]{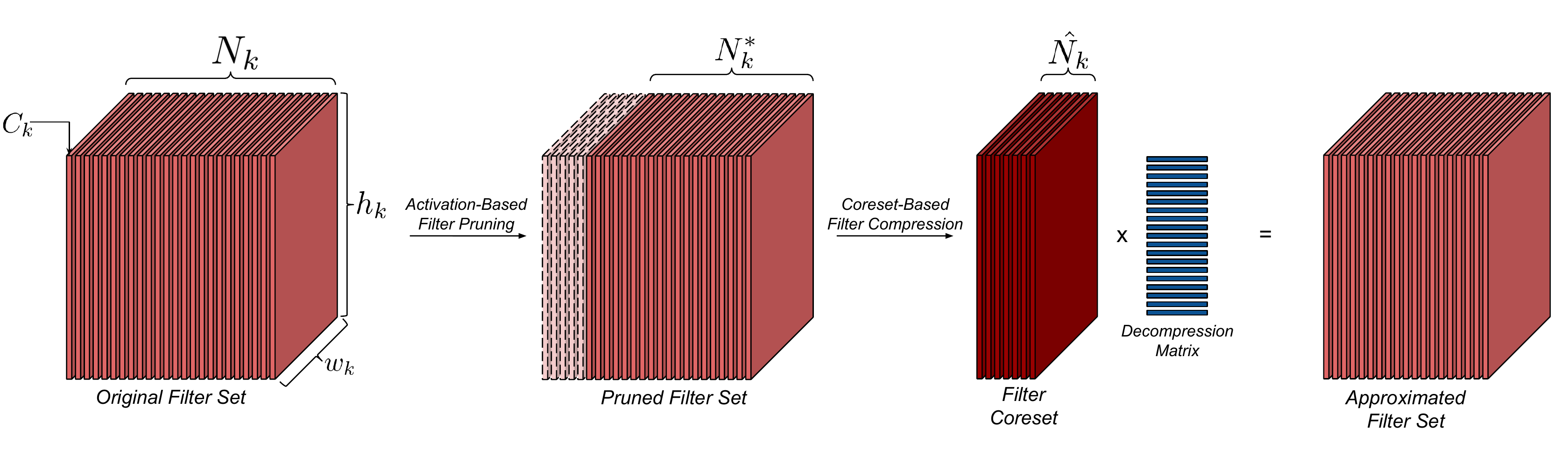}
  \caption{Visual representation of compression pipeline for layer $k$ of a neural network. Our algorithm proceeds in two steps - (i) filter pruning, and (ii) filter compression, as illustrated.}
\end{figure*}

\subsection{Activation-Weighted Coresets}
Our final coreset formulation is obtained by introducing a relative importance score to every filter (based on their activation magnitudes over the training set), while inducing sparsity. However, if we attempt to directly learn a coreset representation by minimizing the reconstruction error over all the training set activations, the resulting optimization problem will be difficult to solve, owing to the large size of the activation matrix and its degenerate nature. We thus employ an alternate formulation: for each filter $f$ in a layer, we compute its `importance' $i^{(f)}_k$ as the mean value of its  activation over all training set points, normalized over all filters, in the $k^{th}$ layer. This is given by the following:
\begin{gather}
  i^{(f)}_k = \frac{\bar{\tensorsym A}^{(f)}_k}{\sum_{p=1}^{N_k} \bar{\tensorsym A}^{(p)}_k }; \qquad \text{where \quad}
  \bar{\tensorsym A}^{(f)}_k = \frac{1}{T} \sum_{j=1}^T \lVert \tensorsym A^{(f)}_{k} (j) \rVert_F
\end{gather}
Here, $\tensorsym A^{(f)}_{k} (j)$ is the activation of the $f^{th}$ filter of layer $k$, for training sample $j$, and $T$ denotes the total number of training samples. We then construct the \textit{Importance Matrix} $\matrixsym I_k$ for the layer $k$ by tiling the column vectors $(i^{(f)}_k)_{f=1}^{N_k}$, for $(C_k \times h_k \times w_k + 1)$ times, creating an \textit{Importance Matrix} $\in \mathbb R^{N_k \times (C_k h_k w_k + 1)}$, where each row denotes the `importance' of each filter, normalized over all filters of the current layer.

We create this form of the importance matrix with every element of a row containing identical values, since we do not want to weigh each component of a particular filter differently. Note, additionally, that we can compute this matrix in only one forward pass of the entire training set. This leads us to the following optimization problem:
\begin{gather}
  \min_{\matrixsym U'_k, \matrixsym \Sigma'_k, \matrixsym V'_k} \lVert \matrixsym I_k \odot (\matrixsym W_k - \matrixsym U'_k\matrixsym \Sigma'_k\matrixsym V'^T_k) \rVert_F^2
\end{gather}
Here $\odot$ denotes the Hadamard (elementwise) product. This problem is essentially, a weighted low-rank decomposition, studied previously by Srerbo and Jaakkola~\cite{srebro2003weighted} and Delchambre~\cite{delchambre2014weighted} and is solved using an efficient Expectation-Maximization (EM) algorithm~\cite{srebro2003weighted}.

The intuition behind this weighted formulation is to ascribe a relative importance to the filters that contribute most to the activations in the training set (on average in the Frobenius norm sense), instead of attempting to reconstruct all activations with equal priority. Molchanov~\textit{et al.}~\cite{molchanov2016pruning} also use the notion of an importance criteria for compression but rather than using it as a weighting scheme in the optimization objective, like we do, they directly use it to prune the `less important' filters. In this case as well, we compute the optimum number of components to be kept by selecting the least number of components that can be selected such that the classification accuracy is bounded within 0.5\% of the original network, once the entire network has been compressed.

\subsection{Activation-Based Filter Pruning}
\label{sec:pruning}
In related work, Li\textit{et al.} observe that not all filters are equally important in the context of classification \cite{li2016pruning}. This motivates us to perform a pre-processing step before coreset compression, to first eliminate unimportant filters pre-emptively, based on the mean of their activation norms over the training set. This step is essential to remove unimportant weights, since pruning out a filter in a layer can completely remove the weights corresponding to that filter, in the next layer as well, inducing greater sparsity. Using the notation from earlier we can write the size of filters $\hat{\matrixsym W_k}$ as:
\begin{gather*}
  \text{size}(\hat{\matrixsym W_k}) =  \hat{N_k} \times (C_k h_k w_k + 1) + N_k \hat{N_k}\\ \intertext{Setting $C_k = N_{k-1}$ (since number of outgoing activations in the previous layer is equal to the number of input channels in the next layer), and using $N_{k-1} h_k w_k >> N_k$, we get:}
  \text{size}(\hat{\matrixsym W_k})  \propto \hat{N}_k \cdot N_{k-1}
\end{gather*}
By layer-wise pruning of complete filters, we can hence set the number of post-pruning filters at layer $k-1$ to be $N^*_{k-1} < N_{k-1}$, permitting further compression. In networks with skip-connections (e.g. ResNets), $C_k \neq N_{k-1}$, but it is a positive linear combination of the number of filters of the ``source'' layers of the (skip) connections, hence the proportionality still holds.

Starting from the first layer in the network, we proceed to evaluate the activation values for the entire training set, layer-by-layer. Inspired by standard ``Max-Pool'' sub-sampling techniques prevalent in modern CNNs~\cite{krizhevsky2012imagenet,simonyan2014very}, we approximate the response from each filter in the convolution layers (a 2D matrix) with its maximum value (a scalar). Once we have this set of pooled filter-wise activations for all samples, we compute the mean squared norm of each filter over all the training samples, and sort the filters by this value. This technique of ordering filters differentiates us from prior pruning-based techniques. We maximize the number of pruned filters, ensuring that the divergence in classification accuracy is only  0.5\%, after the pruning has been carried out across all layers. Once we obtain a reduced set of filters with the crucial filters preserved, we compress this set of filters using coresets, as discussed earlier.

\subsection{Compression Pipeline and Computational Complexity Analysis}
The entire pipeline for compression can be summarized in two stages - (i) activation-based pruning, followed by (ii) coreset-based compression. The pruning procedure can be summarized in the following steps:
\begin{enumerate}
  \item Sort the layers of the network in order of descending parameter size.
  \item For each layer of the sorted network, repeat the following steps:
  \begin{enumerate}
    \item Compute activations for every input in the training set, and store the maximum value for each filter activation (max-pool over spatial dimensions).
    \item Sort the filters in descending order of the mean value of the max-pooled activations, over the entire training set.
    \item Find the smallest number of filters $N^*_k$ that can be retained while performance deviation, post the compression of all layers, is within 0.5\% of original performance - using binary search.
  \end{enumerate}
\end{enumerate}
We can see that the complexity of the individual steps are $\mathcal O(n \log n)$ for the first sorting step, and $\mathcal O(n \cdot (A + N_k \log N_k + A \log N_k))$ for layer-wise activation computation, filter sorting, and binary-search. $A$ denotes the complexity to do one feed-forward operation on the entire training set. Since $A >> N_k >> 1$, the total complexity of the filter pruning is $\mathcal O(n \cdot A \cdot \log \max_k N_k)$, requiring a maximum of $n \log \max_k N_k$ epochs of feed-forward operations, which, for most neural network architectures, we find to be much smaller than the complexity of fine-tuning.

After filter pruning, we proceed to the coreset-based compression stage. This procedure for compression can be summarized in the following steps:

For each layer in the network, starting from the shallowest, do:
  \begin{enumerate}
    \item Compute the complete decomposition according to the coreset formulation used.
    \item Find the minimum number of coreset filters $\hat{N_k}$  that can be retained while performance is within 0.5\% of the network prior to coreset compression, post the compression of all layers - by searching over a random subset of the training data using binary search.
  \end{enumerate}

The complexity for the coreset construction set is $\mathcal O(n \cdot (B + sA \log N^*_k))$, where $B$ is the complexity for the matrix decomposition, and $0 \leq s \leq 1$ is the fraction of random training points used. For our experiments, we set $s = 0.005$. We find that for most networks, $sA\log N^*_k > B$, and hence the total complexity of the compression pipeline is~$\mathcal~O(n\cdot~A\cdot~\log~N_k\cdot(1+s))$. Note that post the cascading of activation-based pruning with coreset compression across all layers, the total deviation allowed in classification performance is 1\% (0.5\% for pruning and 0.5\% for coreset compression).
\section{Experimental Evaluation}
We implement our method in PyTorch~\cite{paszke2017pytorch} and Caffe~\cite{jia2014caffe}, and evaluate on a cluster of NVIDIA TITAN Xp and Tesla GPUs. All of our implementation and other details are available here~\footnote{\url{https://sites.google.com/site/metrosmiles/research/research-projects/compress_cnn}}. For all experiments, we evaluate all 3 coreset construction techniques, as well as the impact of activation-based pruning coupled with each, and report all results together with the baseline and comparable recent work. The Activation-Based Pruning pipeline is reported as AP, while the coreset techniques are reported as - (1) k-Means Coreset (Coreset-K), (2) Structured Sparse Coreset (Coreset-S) and (3) Activation-Weighted Coreset (Coreset-A). We compare our compression performance with recent compression benchmarks, such as Fast-Food~\cite{yang2015deep}, SVD~\cite{denton2014exploiting}, Weight-Based Pruning~\cite{han2015learning}, Deep Compression~\cite{han2015deep}, memory-bounded CNNs~\cite{collins2014memory}, Compresssion Aware Training \cite{alvarez2017compression}, etc. on a wide array of CNN architectures, including the highly efficient SqueezeNet~\cite{iandola2016squeezenet}.


\subsection{LeNet-5 on MNIST}
The first architecture we evaluate is the LeNet-5 network~\cite{lecun1998gradient} on the MNIST dataset~\cite{lecun1998mnist}. This is a popular benchmark for network compression, and high values of compression are reported by various recent work, which makes it a very competitive setup. The results for this experiment are summarized in Table~\ref{tab:lenet}. We can see that the coreset-based methods outperform the recent work comfortably, with a relative improvement of 18\% over the existing state-of-the-art.

\subsection{Large-Scale ImageNet Models}
The next set of experiments we perform are on the large-scale ImageNet-trained models - the very deep networks such as Residual Networks~\cite{he2015deep}, AlexNet~\cite{krizhevsky2012imagenet} and VGGNet-16~\cite{simonyan2014very}. These architectures are ubiquitious for countless applied computer vision tasks~\cite{he2017mask,dai2017deformable}, and several recent compression techniques demonstrate remarkable compression on these models which makes them an appropriate benchmark for evaluating compression performance. For these networks, we also demonstrate the impact of coupling \textit{Deep Compression} (which involves quantization, pruning, re-training iteratively) with our method.

\begin{table*}[t]
\scriptsize
\centering
\caption{Compression (Comp.) results for both AlexNet~\cite{krizhevsky2012imagenet} and VGGNet-16~\cite{simonyan2014very} trained on the ImageNet dataset, along with variation in performance with Deep Compression.\label{tab:alex_vgg}}
\begin{tabular}{l|c|c|c|c|c|c|c}
\hline \hline
\multirow{2}{*}{\textbf{Method}} & \multicolumn{4}{c|}{\textbf{AlexNet}~\cite{krizhevsky2012imagenet}} & \multicolumn{3}{c}{\textbf{VGGNet-16}~\cite{simonyan2014very}} \\ \cline{2-8}
& Acc.(\%) & \#Params & Comp. & \#Epochs & Acc.(\%) & \#Params & Comp. \\ \hline
Baseline & 57.22 & 61M & 1$\times$ & - & 68.88 & 138M & 1$\times$ \\
Fastfood-32-AD~\cite{yang2015deep} & 58.07 & 30M & 2$\times$ & - & - & - & - \\
Fastfood-16-AD~\cite{yang2015deep} & 57.10 & 17M & 3.7$\times$ & - & - & - & - \\
Collins \& Kohli~\cite{collins2014memory} & 55.60 & 15.3M & 4$\times$ & - & - & - & - \\
Compression-Aware~\cite{alvarez2017compression} & - & - & - & - & 67.6 & 64.17M & 2.2$\times$ \\ 
SVD~\cite{denton2014exploiting} & 55.98 & 12.2M & 5$\times$ & 540 & 68.85 & 27M & 5.1$\times$ \\
Pruning~\cite{han2015deep} & 57.23 & 6.8M & 9$\times$ & 960 & 68.15 & 15M & 9.1$\times$ \\
Dynamic Net Surgery~\cite{guo2016dynamic} & 56.91 & 3.47M & 17.7$\times$ & 140 & - & - & - \\ \hline
Coreset-K & 56.97 & 9.15M & 6.7$\times$ & 17 & 68.69 & 15.6M & 9.2$\times$ \\
Coreset-S & 56.78 & 5.76M & 10.5$\times$ & 21 & 68.65 & 9.9M & 13.9$\times$ \\
Coreset-A & 56.82 & 4.97M & 12.3$\times$ & 23 & 68.01 & 9.2M & 15.1$\times$ \\ \hline
AP+Coreset-K & 56.51 & 4.02M & 15.2$\times$& 26 & 68.56 & 9.81M & 14$\times$ \\
AP+Coreset-S & 56.38 & \textbf{3.20M} & \textbf{19.1$\times$} & 28 & 67.90 & \textbf{8.1M} & \textbf{17$\times$} \\
AP+Coreset-A & 56.48 & 3.68M & 16.5$\times$ & 27 & 68.16 & 8.7M & 15.8$\times$ \\ \hline
\multicolumn{7}{c}{\textit{With Deep Compression (Comparison of Model Size)}} \\ \hline
Baseline & 57.22 & 6.9MB & 35$\times$&- & 68.70 & 10.77MB & 49$\times$ \\ \hline
Coreset-K & 56.80 & 4.17MB & 49$\times$&- & 68.51 & 2.52MB & 210$\times$ \\
Coreset-S & 56.87 & 3.92MB & 52$\times$&- & 68.25 & 2.35MB & 225$\times$ \\
Coreset-A & 57.19 & 4.01MB & 51$\times$&- & 68.43 & 2.41MB & 220$\times$ \\ \hline
AP+ Coreset-K & 56.85 & 4.01MB & 51$\times$&- & 68.02 & 2.28MB & 232$\times$ \\
AP+ Coreset-S & 56.70 & 3.85MB & 53$\times$&- & 68.16 & 2.26MB & 233$\times$ \\
AP+ Coreset-A & 57.08 & \textbf{3.74MB} & \textbf{55$\times$}&- & 68.14 & \textbf{2.21MB} & \textbf{238$\times$} \\
\hline \hline
\end{tabular}
\end{table*}

Table \ref{tab:resnet} summarizes the empirical evaluation on Residual Networks. We find state-of-the-art performance achieved by all three coreset methods, and a substantial increase from previous baselines as well. Even in 101-layer deep networks such as ResNet-101, we are able to obtain consistent compression, similar to the shallower ResNets. Note that this improvement is entirely on convolutional layers, which typically have very few redundancies when compared to fully-connected layers. We additionally observe that activation-based pruning buys us significant compression, providing in essence a cascading additive effect.

Table \ref{tab:alex_vgg} summarizes the empirical evaluation on AlexNet and VGGNet-16 networks, the two of the largest image classification networks in use today. We demonstrate substantial improvements over the state-of-the-art, by compressing AlexNet by \textbf{19$\times$}, and VGGNet-16 by \textbf{17$\times$} from their baseline sizes. When combined with Deep Compression, these ratios increase, up to $55\times$ and $238\times$ respectively, yielding models with a memory footprint of less than 4MB. The results additionally highlight the improvement that the activation-based pruning (AP) provides, which is most prominent in the Coreset-K and Coreset-S models.

\begin{table}[!htb]
\begin{minipage}{.38\linewidth}
  \centering\scriptsize
\caption{Compression (Comp.) results on LeNet-5.\label{tab:lenet}}
  \begin{tabular}{l|c|c}
    \hline \hline
    \textbf{Method} & \textbf{Top-1} & \textbf{Comp.} \\ \hline
    Baseline & 0.97 & 1$\times$ \\ \hline
    Wang~\textit{et al.}\cite{wang2016training} & 0.93 & 16$\times$\\
    Han~\textit{et al.}\cite{han2015deep} & 0.74 & 39$\times$ \\
    Guo~\textit{et al.}\cite{guo2016dynamic} & 0.91 & 108$\times$ \\
    SVD~\cite{denton2014exploiting} & 0.92 & 118$\times$ \\
    Ullric,~\textit{et al.}\cite{ullrich2017soft} & 0.97 & 164$\times$ \\\hline
    AP+Coreset-K & 0.966 & 165$\times$ \\
    AP+Coreset-S & 0.96 & 192$\times$ \\
    AP+Coreset-A & 0.96 & \textbf{193}$\times$ \\ \hline \hline

   \end{tabular}
\end{minipage}
\hspace{.02\linewidth}
\begin{minipage}{.57\linewidth}
  \centering\scriptsize
\caption{Compression results on Residual Networks. Columns Acc. and Comp. represent the Top-1 accuracy and compression factor respectively.\label{tab:resnet}}
  \begin{tabular}{l|c|c|c|c|c|c}
    \hline \hline
    \multirow{3}{*}{\textbf{Method}} & \multicolumn{6}{c}{\textbf{Residual Network}} \\ \cline{2-7}
    & \multicolumn{2}{c|}{\textbf{Res-18}} & \multicolumn{2}{c|}{\textbf{Res-50}} & \multicolumn{2}{c}{\textbf{Res-101}} \\ \cline{2-7}
    & Acc. & Comp. & Acc. & Comp. & Acc. & Comp. \\ \hline
    Baseline~\cite{he2016deep} & 0.69 & 1$\times$ & 0.75 & 1$\times$ & 0.76 & 1$\times$ \\
    SVD~\cite{denton2014exploiting} & 0.69 & 8$\times$ & 0.74 & 9.1$\times$ & 0.75 & 9.2$\times$ \\
    Pruning~\cite{han2015learning} & 0.68 & 5.2$\times$ & 0.74 & 6.2$\times$ & 0.76 & $6.4\times$ \\
    N2N~\cite{ashok2017n2n} & 0.67 & 9.0$\times$ & 0.73 & 8.7$\times$ & 0.74 & 8.5$\times$
    \\
    ThiNet~\cite{luo2017thinet} & - & - & 0.71 & 2.06$\times$ & - & - \\
    ThiNet~\cite{luo2017thinet} & - & - & 0.68 & 2.95$\times$ & - & - \\ \hline
    AP+Coreset-K & 0.69 & 13.3$\times$ & 0.74 & 14.7$\times$ & 0.75 & 15.1$\times$\\
    AP+Coreset-S & 0.68 & \textbf{15}$\times$ & 0.74 & \textbf{15.8}$\times$ & 0.75 & \textbf{16.2}$\times$\\
    AP+Coreset-A & 0.69 & 14.2$\times$ & 0.74 & 15.6$\times$ & 0.75 & 15.8$\times$ \\ \hline \hline

   \end{tabular}
\end{minipage}
\end{table}
\begin{table}[t]
\centering\scriptsize
\caption{Comparison with SqueezeNet~\cite{iandola2016squeezenet} trained on the ImageNet dataset. We can compress SqueezeNet to create a model that is \textbf{832}$\times$ smaller than AlexNet~\cite{krizhevsky2012imagenet} with the same performance.\label{tab:squeezenet}}
\begin{tabular}{l|c|c|c|c}
\hline \hline
\multirow{2}{*}{\textbf{Method}} & \multirow{2}{*}{\textbf{Acc.(\%)}} & \textbf{Num. of} & \multirow{2}{*}{\textbf{Ratio}} & \textbf{Rel. to} \\
& & \textbf{Params} & & \textbf{AlexNet} \\ \hline
Baseline & 57.01 & 1.24M & 1$\times$ & 50$\times$ \\ \hline
Coreset-K & 56.83 & 0.73M & 1.7$\times$ & 85$\times$ \\
Coreset-S & 56.92 & 0.65M & 1.9$\times$ & 95$\times$ \\
Coreset-A & 56.94 & 0.61M & 2$\times$ & 102$\times$ \\ \hline
AP+ Coreset-K & 56.52 & 0.65M & 1.9$\times$ & 95$\times$ \\
AP+ Coreset-S & 56.44 & 0.59M & \textbf{2.1$\times$} & \textbf{109$\times$} \\
AP+ Coreset-A & 56.80 & 0.60M  & 2$\times$ & 103$\times$ \\ \hline
\multicolumn{5}{c}{\textit{With Deep Compression (Comparing Model Size)}} \\ \hline
Baseline & 56.04 & 0.47MB & 10.14$\times$ & 507$\times$ \\ \hline
Coreset-K & 56.08 & 0.29MB & 16.1$\times$ & 805$\times$ \\
Coreset-S & 56.05 & 0.28MB & 16.34$\times$ & 817$\times$ \\
Coreset-A & 56.03 & 0.29MB & 16.23$\times$ & 812$\times$ \\ \hline
AP+ Coreset-K & 56.31 & 0.27MB & 16.50$\times$ & 825$\times$ \\
AP+ Coreset-S & 56.15 & 0.26MB & \textbf{16.64$\times$} & \textbf{832$\times$} \\
AP+ Coreset-A & 56.18 & 0.27MB  & 16.56$\times$ & 828$\times$ \\
\hline \hline
\end{tabular}
\end{table}

\subsection{SqueezeNet}
We evaluate our method on the highly parameter-efficient SqueezeNet architecture to evaluate if further redundancies still persist after such a compression in the architecture space and if those can be eliminated via efficient filter bank representations. We find that despite beginning with $50\times$ less parameters than AlexNet (while providing the same performance), SqueezeNet can be compressed further (results in Table ~\ref{tab:squeezenet}). Using our method, we are able to compress SqueezeNet to half its parameters, providing accuracy similar to AlexNet at $100\times$ compression. By coupling with Deep Compression, we obtain a net compression in model size to the tune of $16.64\times$ over the original model (or $\bf 832\times$ from AlexNet) while maintaining classification performance.

\subsection{Additional Observations}
\begin{table}
  \centering\scriptsize
  \caption{LeNet-5 layer-wise compression of our method (denoted by identifiers) vis-\'a-vis prior work. The entries represent the fraction of parameters retained post compression.
  \label{tab:layer_wise}}
  \begin{tabular}{l|c|c|c|c|c|c|c|c}
    \hline \hline
    \textbf{Layer} & \textbf{Han \textit{et al.}}\cite{han2015learning} & \textbf{Guo \textit{et al.}}\cite{guo2016dynamic} & \textbf{K}    & \textbf{S}    &  \textbf{A}   & \textbf{AP+K} & \textbf{AP+S} & \textbf{AP+A} \\ \hline
    conv1 & 0.66 & 0.14 & 0.06 & 0.03 & 0.03 & 0.02 & \textbf{0.02} & 0.02 \\
    conv2 & 0.12 & 0.03 & 0.04 & 0.03 & 0.03 & 0.02 & \textbf{0.02} & 0.02\\
    fc1 & 0.08 & \textbf{0.01} & 0.04 & 0.03 & 0.03 & 0.02 & \textbf{0.01} & 0.02 \\
    fc2   & 0.19 & 0.04 & 0.02 & 0.01 & 0.02 & 0.01 & \textbf{0.01} & 0.01 \\
    \hline \hline
  \end{tabular}
\end{table}

Further, we observe that Coreset-S and Coreset-A formulations consistently outperform Coreset-K. We surmise that large extant model redundancies tend to benefit Coreset-A and S formulations where sparsity is explicitly enforced in the objective. Moreover, we observe that for deeper models Coreset-S tends to achieve the most compression. Table~\ref{tab:layer_wise} shows the superior layer-wise compression achieved by our algorithm vis-\'a-vis state-of-the-art compression techniques on LeNet-5. The results clearly bring out the efficacy of using our compression technique, especially for convolution layers. For layer-wise compression results on other CNNs, please refer to the supplementary.

\textbf{Runtime Analysis:} We also perform a study of runtime analysis in both training and inference performance. Since we do not undertake retraining, our method is considerably faster - on our hardware, one forward pass and backward pass of AlexNet (batch size 256) takes 16ms naively, which corresponds to a total epoch training time (on ImageNet) of 2.5 minutes. We use this as a base measurement to compare the \textbf{total} training time (inclusive of the coreset operations). Table~\ref{tab:alex_vgg} describes the comparison of training times across methods. The previous state of the art method, Dynamic Net Surgery~\cite{guo2016dynamic}, requires 140 epochs (in time units) whereas our method takes \textit{at most} 28 epochs (in time units), a significant reduction of 80\%. During inference, we observe a reduction in inference time as well, which can be optimized by using efficient tensor multiplication~\cite{solomonik2014provably}. On ResNet-50, VGGNet-16 and AlexNet, the naive (uncompressed) runtimes per epoch are: 36ms, 45ms and 8ms respectively. Our best runtimes for these networks (with Coreset-S) are 19ms, 21ms and 3.5ms on average, which is an average improvement of around 50\%.
\begin{figure}[t]
 \begin{minipage}[t]{.3\linewidth}
  \centering
  \includegraphics[width=\textwidth]{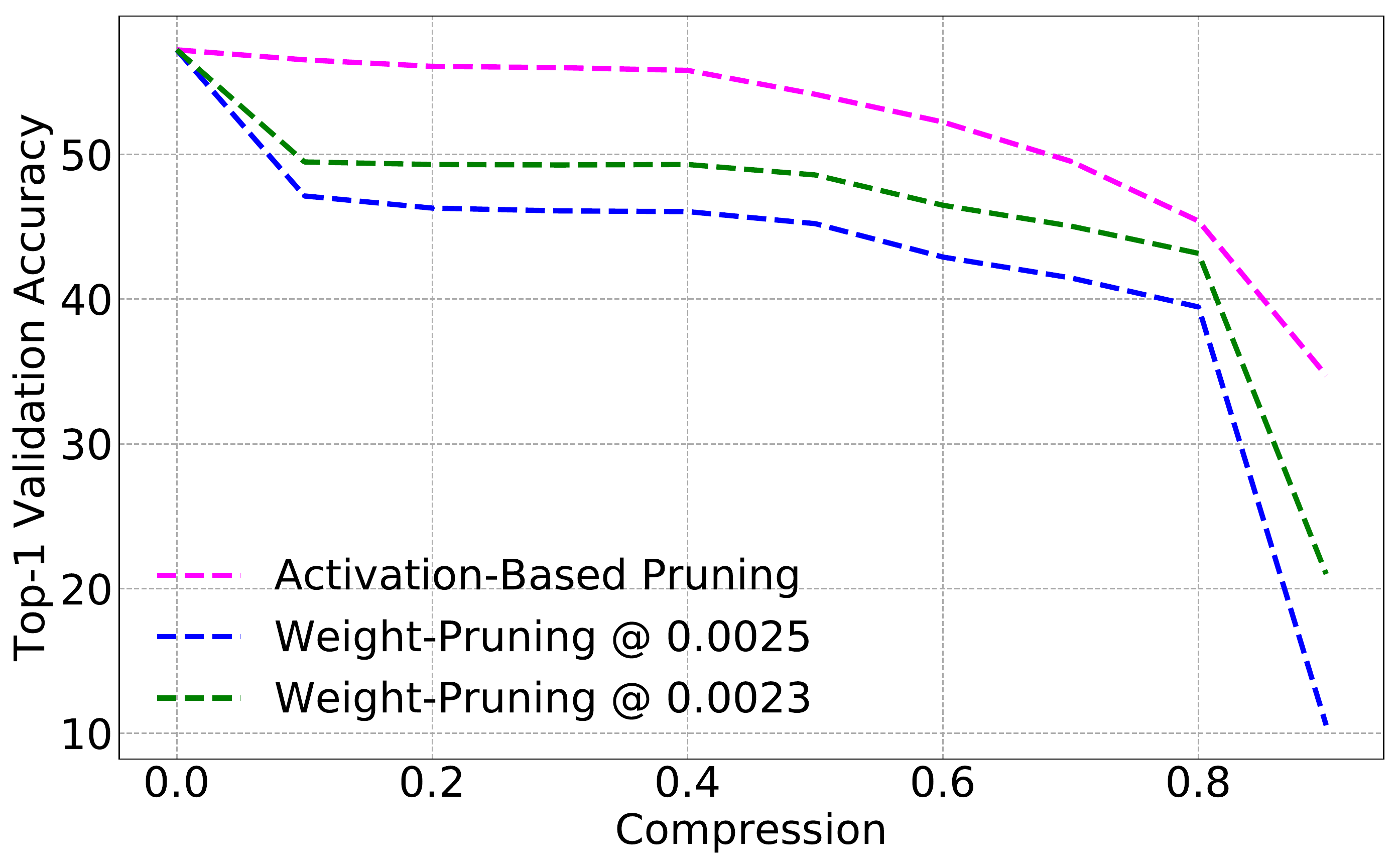}
  \caption{The comparison of Activation-Based Pruning (AP) with weight-based filter pruning (without re-training), on AlexNet.}
  \label{fig:ablation}
  \end{minipage}
  \hspace{.01\linewidth}
  \begin{minipage}[t]{.3\linewidth}
  \centering
  \includegraphics[width=\textwidth]{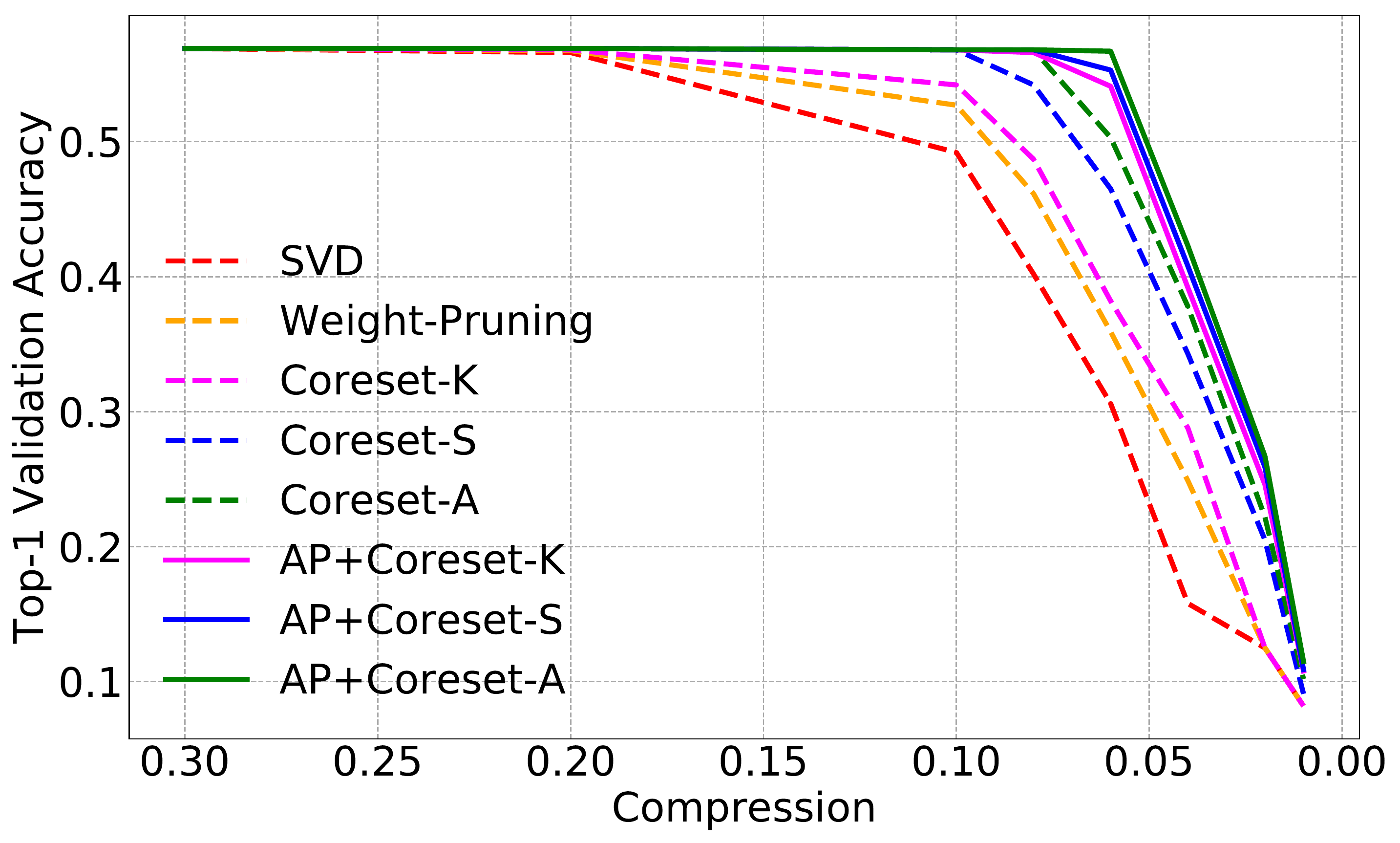}
  \caption{Variation of classification performance with compression for all coreset compression techniques evaluated on AlexNet~\cite{krizhevsky2012imagenet}.}
  \label{fig:variation}
  \end{minipage}
  \hspace{.01\linewidth}
  \begin{minipage}[t]{.3\linewidth}
    \centering
    \includegraphics[width=\textwidth]{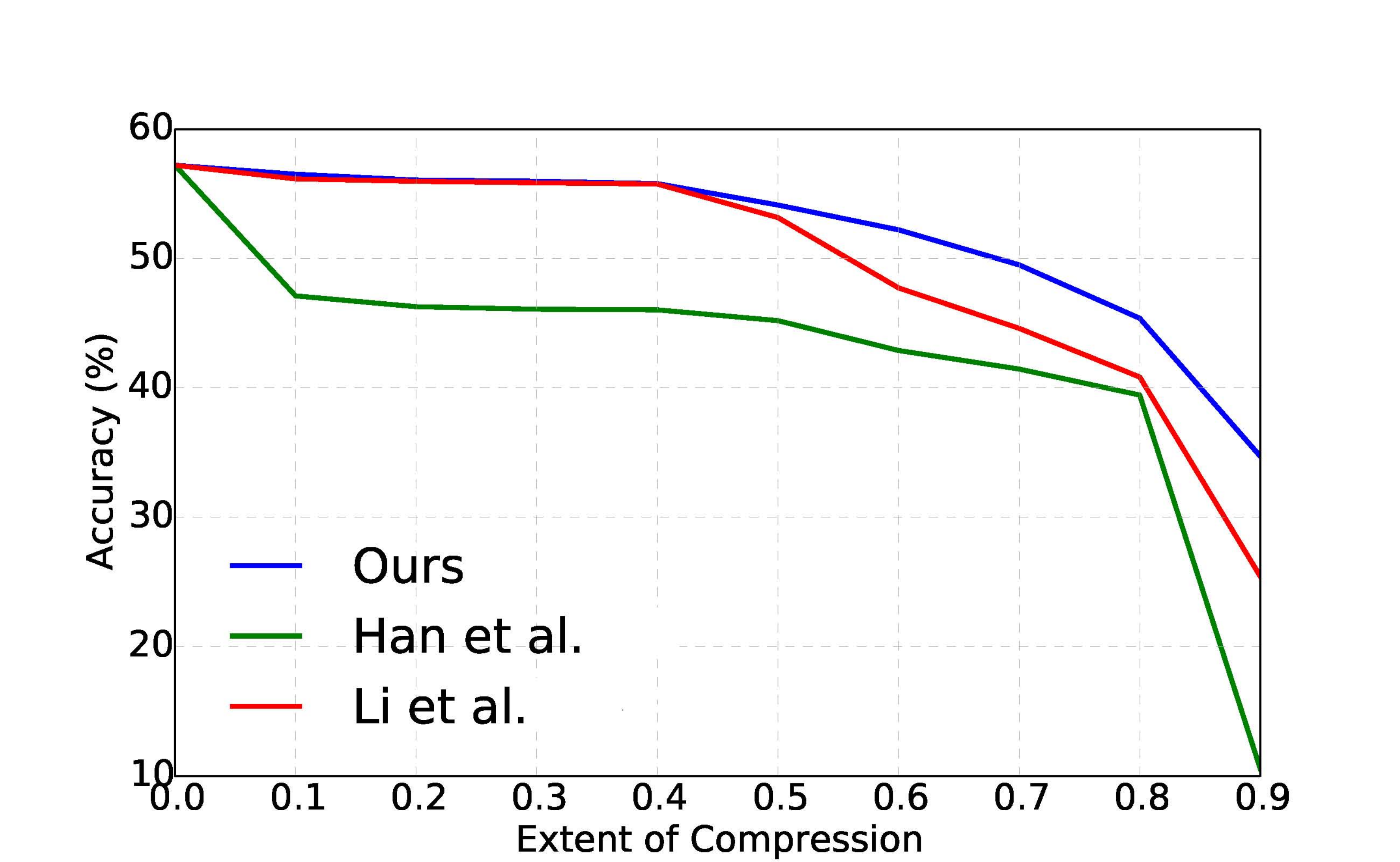}
    \caption{The comparison of Activation-Based Pruning (AP) with the pruning techniques of Han~\textit{et al.} and Li~\textit{et al.}~\cite{han2015learning,li2016pruning}, on AlexNet.}
    \label{fig:ablation2}
  \end{minipage}
\end{figure}

\subsection{Ablation Analysis}
To demonstrate the effect of individual components in our method, we perform some ablation studies as well. We first compare the effect of activation-based pruning (AP) on all coreset compression techniques on three models - AlexNet\cite{krizhevsky2012imagenet}, VGGNet-16\cite{simonyan2014very} and SqueezeNet\cite{iandola2016squeezenet}, and observe that pruning benefits all methods of coreset compression, as described in Tables \ref{tab:alex_vgg} and \ref{tab:squeezenet}.

Next, we compare activation-based pruning with weight-based pruning, without re-training, and the pruning technique of Li~\textit{et al.}~\cite{li2016pruning} for AlexNet. The results of these comparisons are summarized in Figures~\ref{fig:ablation}. We obtain consistently better performance at all compression ratios, substantiating the merit of data-dependent filter pruning approaches over those based on the magnitudes of filter weights.

Finally, we analyze the variation of performance with compression factor for all coreset compression techniques on the AlexNet classification model, described in Figure~\ref{fig:variation}. We observe that Coreset-K (with and without AP), while stronger than SVD and Pruning approaches, worsens much more rapidly in comparison to other corresponding coreset techniques. This observation is consistent across all models. For additional results, more layer-wise compression analysis, and filter visualizations we refer the reader to the supplementary material.

\begin{table}[t]
  \centering\small
\caption{Performance of coreset-based compression on domain-adaptation tasks.\label{tab:fgvc}}
  \begin{tabular}{l|c|c|c}
    \hline \hline
    \textbf{Dataset} & \textbf{Model} & \textbf{\#Params} & \textbf{Top-1}\\ \hline
    \multirow{6}{*}{CUB-2011} & VGG-16 Finetune(FT) & 138M &  72.30\% \\
    & Train from Scratch & 138M & 44.27\% \\ \cline{2-4}
    & SVD\cite{denton2014exploiting} + FT & 27M & 53.65\% \\
    & Pruning\cite{han2015learning} + FT & 15M & 57.45\% \\
    & AP+Coreset-S + FT & 8.1M & \textbf{70.66\%} \\ \hline
    \multirow{6}{*}{Stanford-Dogs} & VGG-16 Finetune(FT) & 138M &  61.92\% \\
    & Train from Scratch & 138M & 27.16\% \\ \cline{2-4}
    & SVD\cite{denton2014exploiting} + FT & 27M & 40.84\% \\
    & Pruning\cite{han2015learning} + FT & 15M & 43.28\% \\
    & AP+Coreset-S + FT & 8.1M & \textbf{55.91\%} \\ \hline \hline
  \end{tabular}
\end{table}

\subsection{Domain Adaptibility}
To measure the generalizability of our compressed models to newer tasks, we evaluate compressed models on domain adaptation benchmarks, following the experimental pipeline proposed in~\cite{luo2017thinet}. We evaluate the performance of the compressed CNN model VGGNet-16~\cite{simonyan2014very} on target domain adaptation datasets - CUB-2011\cite{wah2011caltech} and Stanford-Dogs~\cite{khosla2011novel}, two popular datasets for fine-grained image classification. These results are summarized in Table~\ref{tab:fgvc}. We observe that our compressed coreset models are able to provide classification performance close to the uncompressed networks, while surpassing networks compressed by other techniques. This exhibits the versatility of coreset-compressed models to domain adaptation tasks, as well.
\section{Conclusions and Future Work}
In this paper we introduce a novel technique that exploits redundancies in the space of convolutional filter weights and sample activations to reduce neural network size, using the long-existing concepts of coresets, coupled with an activation-based pooling technique. The lack of a re-training step in our algorithmic pipeline makes the implementation simple. Empirical evaluation reveals that our algorithm outperforms all other competing methods at compressing a wide array of popular CNN architectures. Our findings uncover the existence of redundancies even in the most compressed CNNs, such as SqueezeNets, which can be further exploited to improve efficiency.


Our method does not require any retraining, scales to both convolution and fully connected layers, and is extensively generalizable to different neural network models without being computationally intensive. Thus, we hope that our algorithm will serve as a valuable tool to obtain leaner and more efficient CNNs. As future work, we hope to apply our algorithm to compress other types of deep neural networks, such as Recurrent Neural Networks (RNNs) which are applicable to time-varying sequential inputs.

\noindent\textbf{Acknowledgments:} We are grateful to Prof. Ramesh Raskar for his insightful comments. MC additionally acknowledges Po-han Huang for helpful discussions and NVIDIA for providing the GPUs used for this research.

\newpage

\title{Coreset-Based Neural Network Compression : Supplementary Material}

\titlerunning{Coreset-Based Neural Network Compression : Supplement}
%
\author{Abhimanyu Dubey\thanks{Equal Contribution.}\inst{1} \and
Moitreya Chatterjee\printfnsymbol{1}\inst{2} \and
Narendra Ahuja\inst{2}}
\authorrunning{A. Dubey$^*$, M. Chatterjee$^*$ and N. Ahuja}
%

\institute{Massachusetts Institute of Technology, Cambridge MA 02139, USA \\
\email{dubeya@mit.edu}\and
University of Illinois at Urbana-Champaign, Champaign IL 61820, USA\\
\email{metro.smiles@gmail.com, } \email{n-ahuja@illinois.edu}}
\maketitle
In this document, we represent the 3 coreset techniques and the activation-based pruning step (AP) of our algorithm as optimization problems. We then present accuracy versus compression plots for the 3 coreset techniques and their counterparts when coupled with AP, for AlexNet, VGGNet-16, ResNet-18, ResNet-50, ResNet-101, SqueezeNet, and LeNet-5. In these plots our proposed algorithms are juxtaposed with two competing state-of-the-art techniques, viz. SVD~\cite{denton2014exploiting}, and Weight-Pruning with retraining~\cite{han2015learning}. This is followed by a tabulation of the layer-wise compression achieved by the 3 coreset techniques and their analogues coupled with AP  for AlexNet, VGGNet-16, and LeNet-5. Finally, we conclude with some conv1 filter visualizations for AlexNet, ResNet-18, ResNet-50, and ResNet-101, showing the change brought about by the use of the Coreset-S compression technique.

\section{Optimization Procedure}

\subsection{Coreset-K}
For the k-Means coreset (Coreset-K), we solve the following optimization procedure.
\begin{equation}
\begin{aligned}
   \min_{ \hat{W}_k} \lVert \matrixsym W_k - \matrixsym{\hat{W}_k} \rVert_F^2  \\
   \text{subject to: } \ \text{rank}(\hat{W}_k) < \text{rank}(W_k)
  \label{eq:ck}
\end{aligned}
\end{equation}
The solution to the above is given by the Singular-Value-Decomposition (SVD) of $W_k$.

\subsection{Coreset-S}
For the sparse coreset (Coreset-S), we solve the following optimization procedure.
\begin{equation}
\begin{aligned}
  \min_{\matrixsym U'_k, \matrixsym \Sigma'_k, \matrixsym V'_k} \lVert \matrixsym W_k - \matrixsym U'_k\matrixsym \Sigma'_k\matrixsym V'^T_k \rVert_F^2 + \lambda \cdot \lVert V'_k \rVert_1 \\
  \text{subject to } \ \lVert (\matrixsym U'_k \cdot \Sigma'_k)_{m} \rVert_2 = 1 \ \forall \ m \in [1, \hat{N_k}]
\end{aligned}
\end{equation}
To solve this optimization, we utilize Algorithm 1 of Jenatton\textit{ et al.}~\cite{jenatton2010structured}, available in standard packages such as \texttt{scikit-learn}. The sparsity values $\lambda$ are obtained via grid-search, and the best $\lambda$ are reported in Table~\ref{tab:lamb}.
\begin{table}
  \centering
  \begin{tabular}{lc}
    \hline \hline
    Network & $\lambda$ \\
    \hline
    AlexNet & 1 \\
    VGGNet & 1.25 \\
    ResNet-18 & 1.25 \\
    ResNet-50 & 1.25 \\
    ResNet-101 & 1.25 \\
    LeNet-5 & 1.5 \\
    SqueezeNet & 1.5 \\ \hline \hline
  \end{tabular}
  \caption{Values of sparsity parameter $\lambda$ obtained by grid-search for Coreset-S compression.}
  \label{tab:lamb}
\end{table}

\subsection{Coreset-A}
For the activation-based coreset (Coreset-A) we solve the following optimization problem:
\begin{gather}
  \min_{\matrixsym U'_k, \matrixsym \Sigma'_k, \matrixsym V'_k} \lVert \matrixsym I_k \odot (\matrixsym W_k - \matrixsym U'_k\matrixsym \Sigma'_k\matrixsym V'^T_k) \rVert_F^2
\end{gather}
Where, the \textit{Importance Matrix}, $\matrixsym I_k$, is specified as:
\begin{gather}
  i^{(f)}_k = \frac{\bar{\tensorsym A}^{(f)}_k}{\sum_{p=1}^{N_k} \bar{\tensorsym A}^{(p)}_k } \\ \intertext{where,}
  \bar{\tensorsym A}^{(f)}_k = \frac{1}{T} \sum_{j=1}^T \lVert \tensorsym A^{(f)}_{k} (j) \rVert_F
\end{gather}
To solve this, we first compute the \textit{Importance Matrix} over the entire training set (using only 1 epoch of forward passes). After that, we solve the above problem in an EM setting as described by Srebro and Jaakkola~\cite{srebro2003weighted} to obtain our decomposition.

\section{Pruning as an Optimization}
The activation-based pruning step of our algorithm denoted by AP, seeks to retain only the top $N_k^*$ filters out of a total of $N_k$ filters in layer k, which have the highest average activation, where $N_k^* < N_k$. In the process of choosing these filters, we make use of the activation response matrix, at layer k, $A_k \in \mathbb{R}^{S \times N_k}$, where S is the number of samples in the training set. Upon pruning, we seek the matrix $\hat{A}_k \in \mathbb{R}^{S \times N_k}$, where $N_k - N_k^*$ columns are fully zeros, representing the fact that $N_k - N_k^*$ filters have been pruned. This may be cast as the following optimization problem:
\begin{equation}
\begin{aligned}
& \underset{\hat{A}_k}{\text{minimize}}
& & || A_k - \hat{A}_k ||_F^2 \\
& \text{subject to}
& & \hat{A}_k = A_k \odot T, \; T \in \{0, 1\}^{S \times N_k}; A_k \in \mathbb{R}^{S \times N_k}, \\
&&& \sum_j \mathbbm{1}_{\{T_{ij} = 1\}} = N_k^*; \forall i \in \{1, 2, ..., S\}, N_k^* < N_k,\\
&&& \sum_i \mathbbm{1}_{\{T_{ij} = 1\}} \in \{0, S\} \forall j \in \{1, 2, ..., N_k\},
\end{aligned}
\end{equation}

where $\mathbbm{1}_{\{\cdot\}}$ indicates the indicator function, which is 1 when the condition $\{\cdot\}$ is true and is 0 else, and $\odot$ indicates the element-wise Hadamard product.
The solution to this optimization is obtained by preserving the top $N_k^*$ columns of $A_k$ ($N_k^* < N_k$), sorted in descending by their average activation values.
\section{Variation of Accuracy with Compression}
In this section we present the plots representing the top-1 accuracy as a function of compression for the 3 different coreset compression algorithms and the 3 AP+coreset compression algorithms for AlexNet, VGG-16, ResNet-18, ResNet-50, ResNet-101, SqueezeNet, and LeNet-5 CNNs. Additionally, we compare our approach with state-of-the-art compression techniques SVD~\cite{denton2014exploiting}, and Weight-Pruning coupled with Retraining~\cite{han2015learning}.

\subsection{AlexNet} 
\begin{figure}
  \centering
  \includegraphics[width=0.48\textwidth]{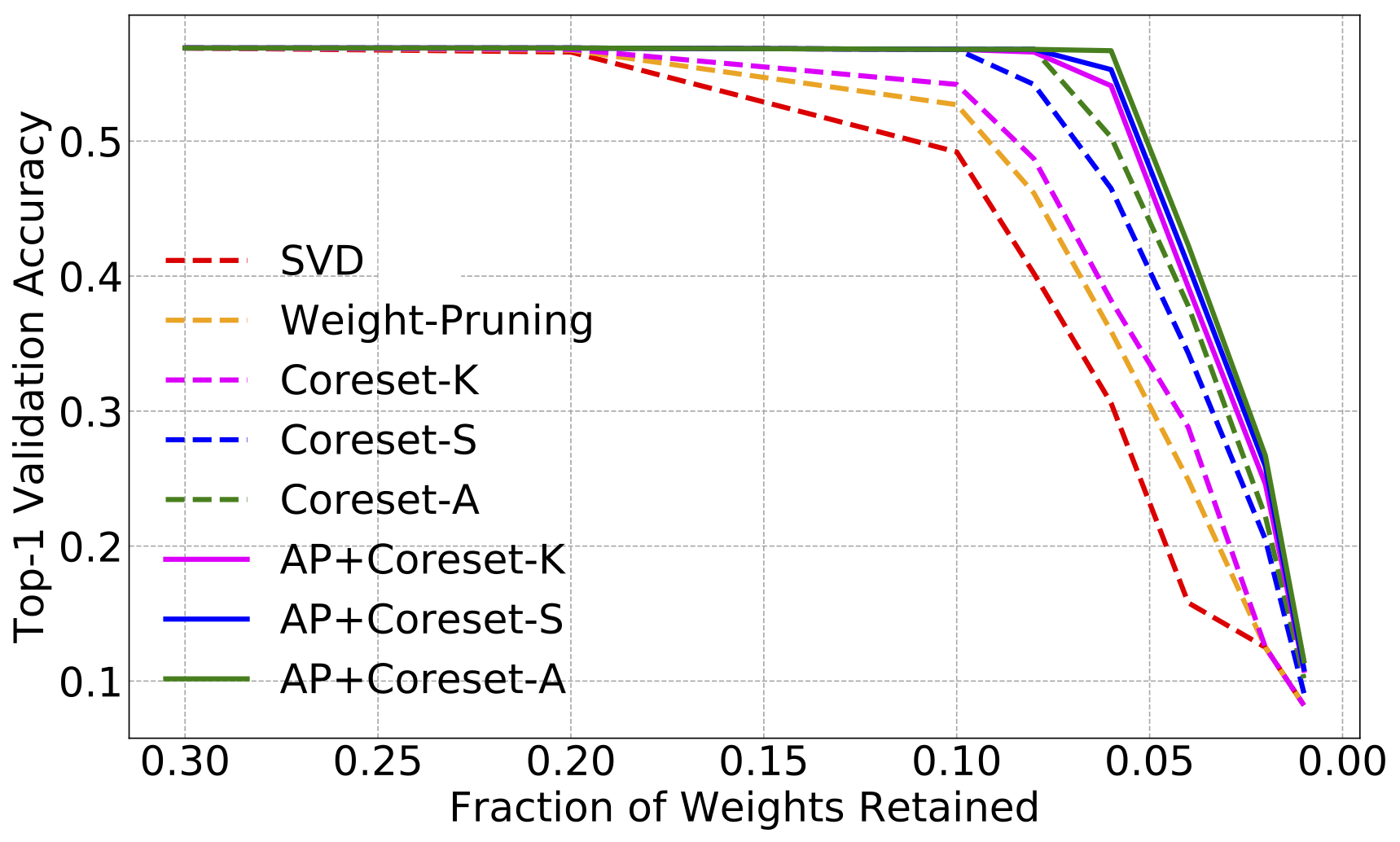}
  \caption{Variation of classification performance with compression on AlexNet.}
  \label{fig:comp1}
\end{figure}

The change in classification performance (accuracy) with variation in fraction of retained model weights for AlexNet is described in Figure~\ref{fig:comp1}.

\subsection{VGGNet-16} 
\begin{figure}
  \centering
  \includegraphics[width=0.48\textwidth]{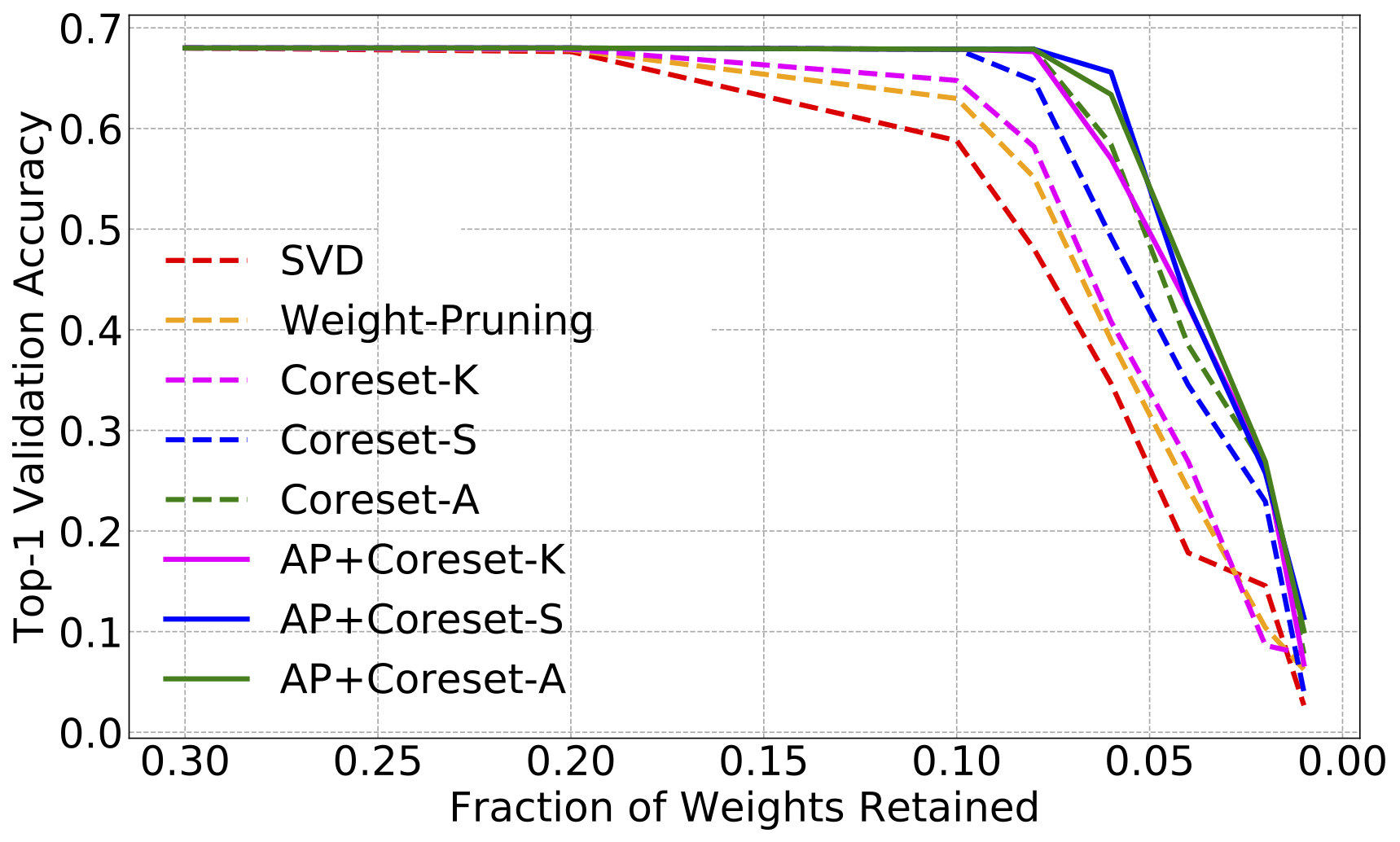}
  \caption{Variation of performance with compression on VGGNet-16.}
  \label{fig:comp2}
\end{figure}

The change in classification performance (accuracy) with fraction of retained model weights for VGGNet-16 is described in Figure~\ref{fig:comp2}.

\subsection{ResNet-18} 
\begin{figure}
  \centering
  \includegraphics[width=0.48\textwidth]{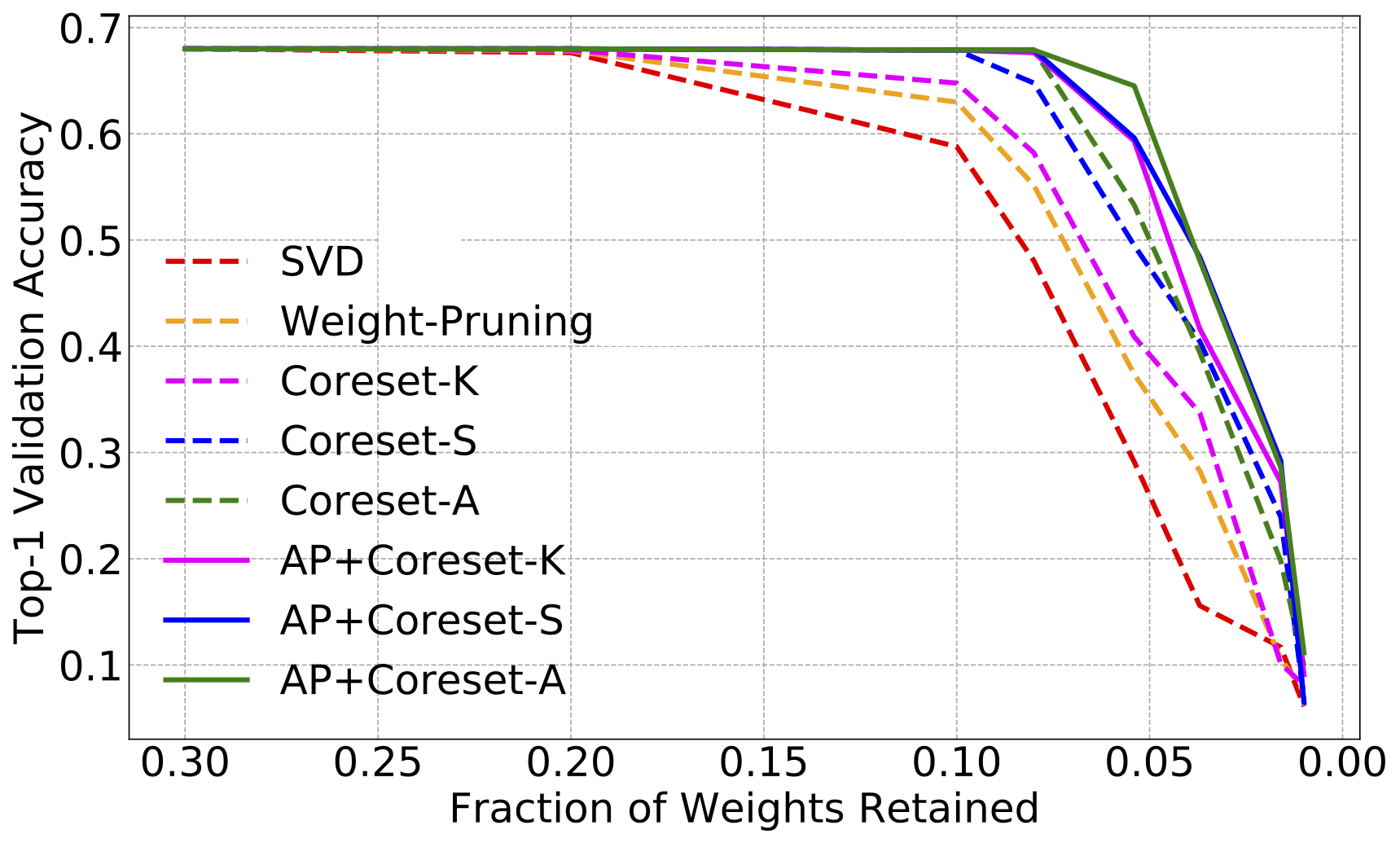}
  \caption{Variation of performance with compression on ResNet-18.}
  \label{fig:comp3}
\end{figure}

The change in classification performance (accuracy) with fraction of retained model weights for ResNet-18 is described in Figure~\ref{fig:comp3}.

\subsection{ResNet-50} 
\begin{figure}
  \centering
  \includegraphics[width=0.48\textwidth]{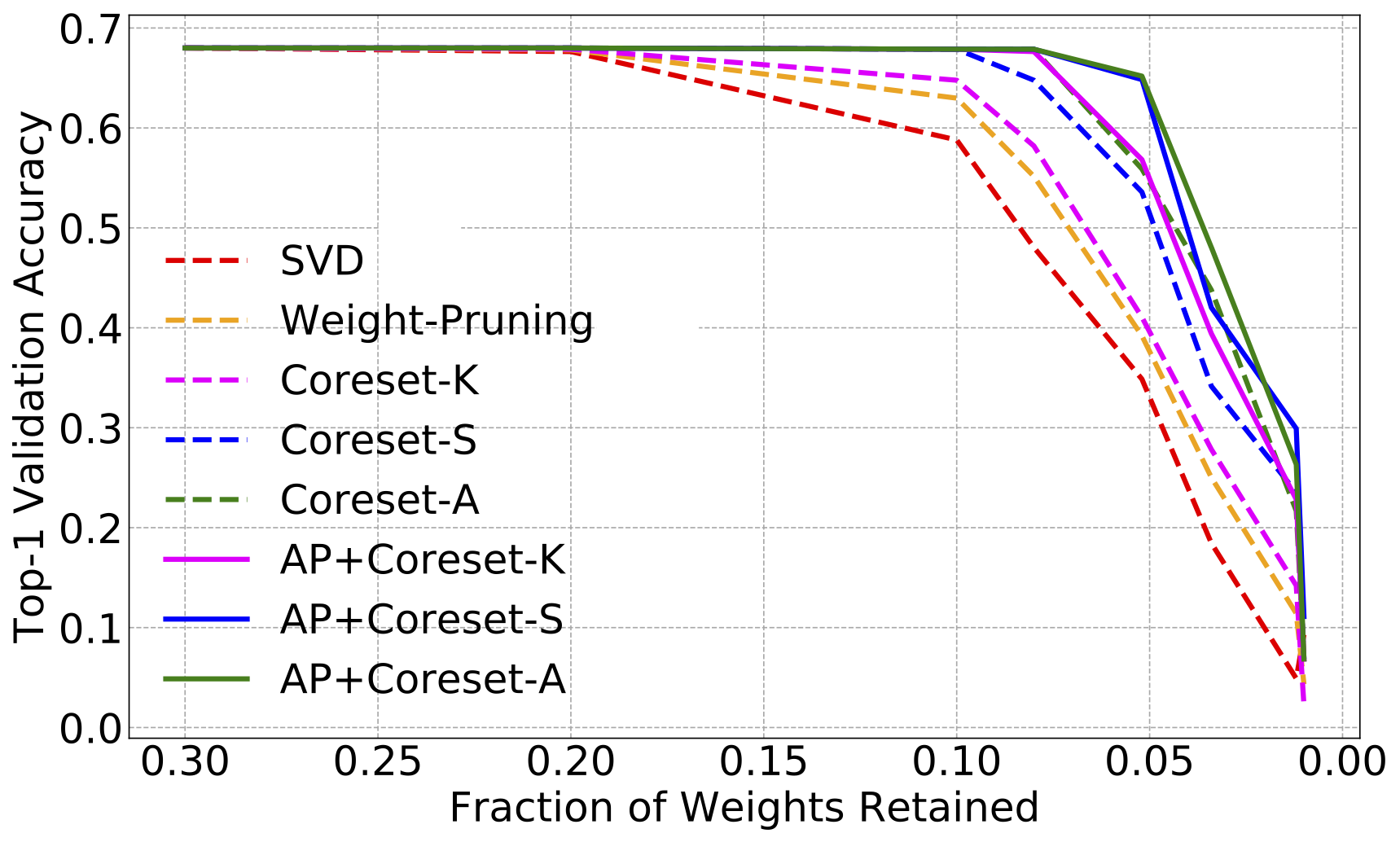}
  \caption{Variation of performance with compression on ResNet-50.}
  \label{fig:comp4}
\end{figure}

The change in classification performance (accuracy) with fraction of retained model weights for ResNet-50 is described in Figure~\ref{fig:comp4}.

\subsection{ResNet-101} 
\begin{figure}
  \centering
  \includegraphics[width=0.48\textwidth]{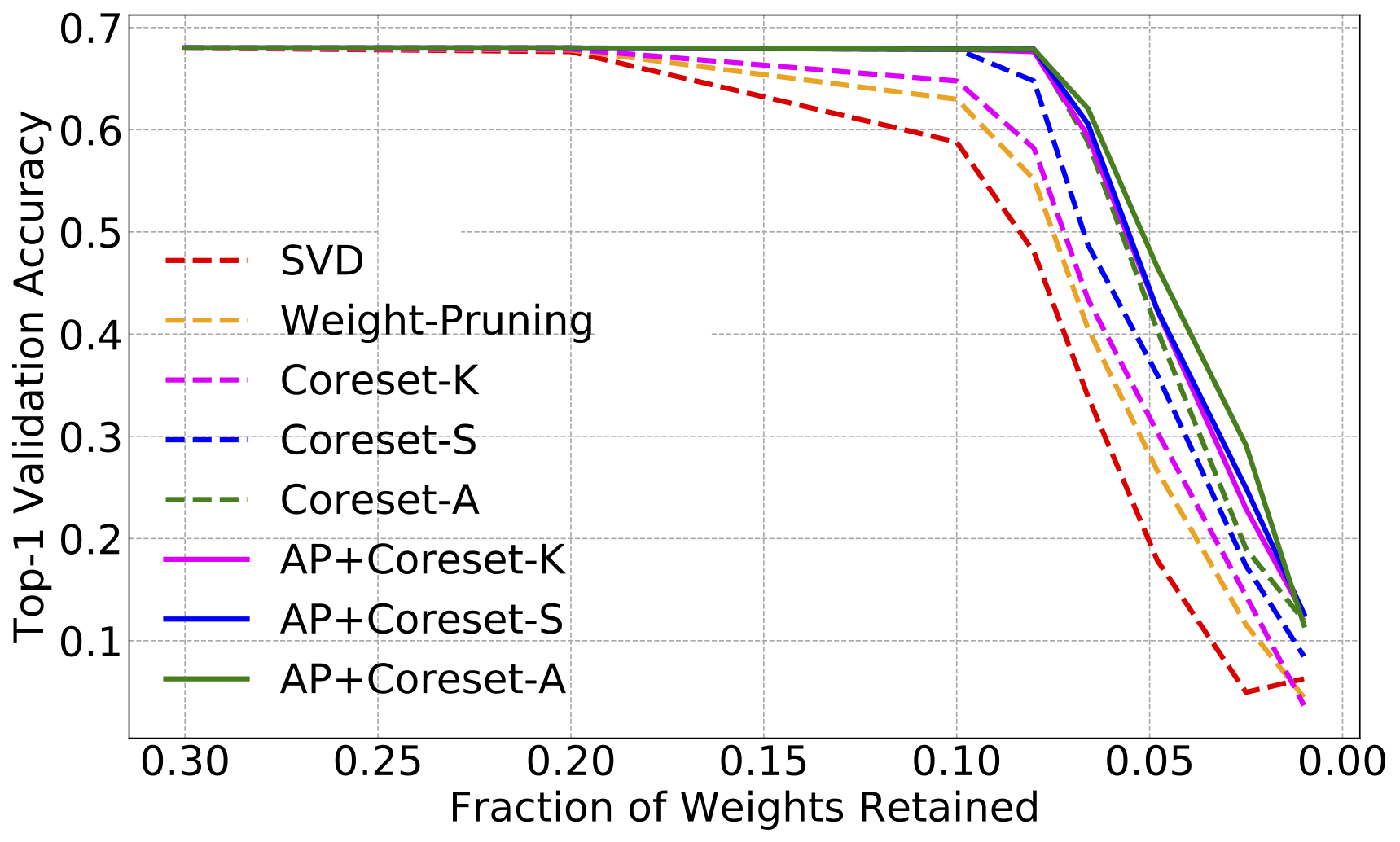}
  \caption{Variation of performance with compression on ResNet-101.}
  \label{fig:comp5}
\end{figure}

The change in classification performance (accuracy) with fraction of retained model weights for ResNet-101 is described in Figure~\ref{fig:comp5}.

\subsection{SqueezeNet} 
\begin{figure}
  \centering
  \includegraphics[width=0.48\textwidth]{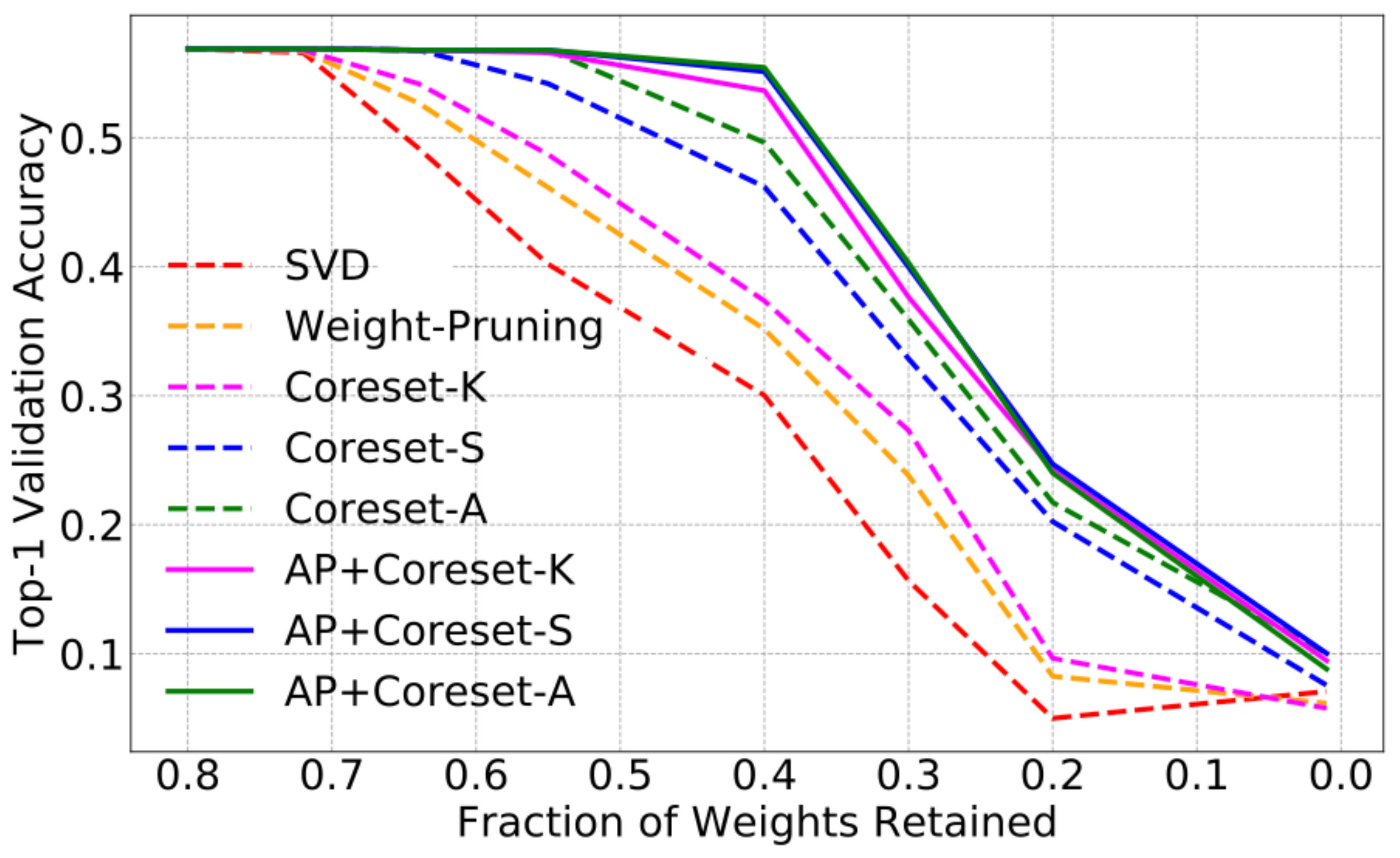}
  \caption{Variation of performance with compression on SqueezeNet.}
  \label{fig:comp6}
\end{figure}

The change in classification performance (accuracy) with fraction of retained model weights for SqueezeNet is described in Figure~\ref{fig:comp6}.

\subsection{LeNet-5} 
\begin{figure}
  \centering
  \includegraphics[width=0.48\textwidth]{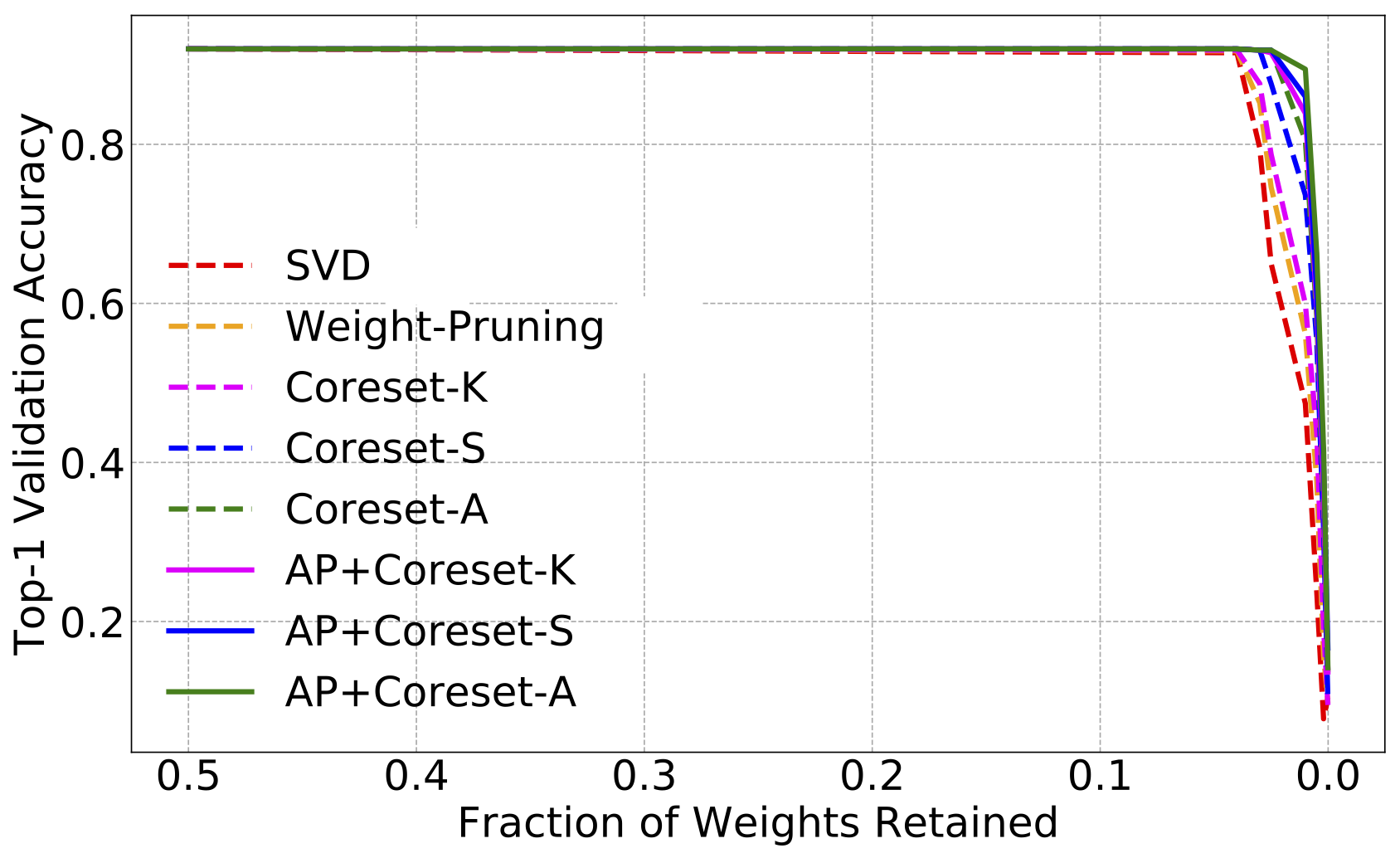}
  \caption{Variation of performance with compression on LeNet-5.}
  \label{fig:comp7}
\end{figure}

The change in classification performance (accuracy) with fraction of retained model weights for LeNet-5 is described in Figure~\ref{fig:comp7}.

\section{Layer-wise Compression}
This section presents tables showing the layer-wise compression for AlexNet, VGGNet-16 and LeNet-5 CNNs, when compressed using the 3 different coreset compression algorithms (Coreset-K (K), Coreset-S (S), Coreset-A (A)), and the 3 AP+coreset compression algorithms.

\subsection{AlexNet}
Table~\ref{tab:sup1} provides layer-wise compression performance for AlexNet over all 6 techniques.
\begin{table}
  \centering
  \begin{tabular}{l||c|c|c|c|c|c|c|c}
    \hline \hline
    \textbf{Layer}  & \textbf{Han \textit{et al.}}\cite{han2015learning} & \textbf{Guo \textit{et al.}}\cite{guo2016dynamic} & \textbf{K}    & \textbf{S}    &  \textbf{A}   & \textbf{AP+K} & \textbf{AP+S} & \textbf{AP+A} \\ \hline
    conv1 & 0.84 & 0.54 & 0.22 & 0.15 & 0.12 & \textbf{0.09} & \textbf{0.09} & \textbf{0.09} \\
    conv2 & 0.38 & 0.41 & 0.21 & 0.14 & 0.14 & \textbf{0.08} & \textbf{0.08} & 0.09 \\
    conv3 & 0.35 & 0.29 & 0.23 & 0.16 & 0.14 & 0.09 & \textbf{0.08} & 0.09 \\
    conv4 & 0.37 & 0.32 & 0.19 & 0.15 & 0.13 & 0.07 & \textbf{0.06} & 0.07 \\
    conv5 & 0.37 & 0.33 & 0.18 & 0.13 & 0.13 & 0.08 & \textbf{0.07} & 0.08 \\
    fc6   & 0.09 & \textbf{0.04} & 0.11 & 0.07 & 0.07 & 0.05 & 0.05 & 0.05 \\
    fc7   & 0.09 & 0.07 & 0.12 & 0.06 & 0.07 & 0.05 & \textbf{0.04} & 0.05 \\
    fc8   & 0.25 & \textbf{0.05} & 0.26 & 0.12 & 0.13 & 0.07 & 0.07 & 0.06\\
    \hline \hline
  \end{tabular}
  \caption{Layer-wise compression for all 6 coreset techniques (denoted only by their identifiers to save space) on AlexNet. The entries represent the fraction of parameters retained post compression.}
  \label{tab:sup1}
\end{table}

\subsection{VGGNet-16}
Table~\ref{tab:sup2} provides layer-wise compression performance for VGGNet-16 over all 6 techniques.
\begin{table}
  \centering
  \begin{tabular}{l||c|c|c|c|c|c|c}
    \hline \hline
    \textbf{Layer} & \textbf{Han \textit{et al.}}\cite{han2015learning} & \textbf{K}    & \textbf{S}    &  \textbf{A}   & \textbf{AP+K} & \textbf{AP+S} & \textbf{AP+A} \\ \hline
    conv1\_1 & 0.58 & 0.21 & 0.14 & 0.11 & 0.08 & \textbf{0.07} & \textbf{0.07} \\
    conv1\_2 & 0.22 & 0.20 & 0.16 & 0.13 & \textbf{0.08} & \textbf{0.08} & \textbf{0.08} \\
    conv2\_1 & 0.34 & 0.23 & 0.15 & 0.15 & \textbf{0.09} & \textbf{0.09} & \textbf{0.09} \\
    conv2\_2 & 0.36 & 0.24 & 0.17 & 0.16 & 0.10 & \textbf{0.08} & 0.10 \\
    conv3\_1 & 0.53 & 0.22 & 0.14 & 0.13 & \textbf{0.08} & \textbf{0.08} & \textbf{0.08} \\
    conv3\_2 & 0.24 & 0.21 & 0.12 & 0.13 & \textbf{0.07} & \textbf{0.07} & 0.08 \\
    conv3\_3 & 0.42 & 0.20 & 0.13 & 0.12 & \textbf{0.07} & \textbf{0.07} & \textbf{0.07} \\
    conv4\_1 & 0.32 & 0.18 & 0.12 & 0.12 & \textbf{0.05} & \textbf{0.05} & \textbf{0.05} \\
    conv4\_2 & 0.27 & 0.18 & 0.12 & 0.13 & \textbf{0.04} & \textbf{0.04} & 0.05 \\
    conv4\_3 & 0.34 & 0.17 & 0.12 & 0.12 & 0.05 & \textbf{0.04} & \textbf{0.04} \\
    conv5\_1 & 0.35 & 0.17 & 0.11 & 0.11 & 0.05 & 0.05 & \textbf{0.04} \\
    conv5\_2 & 0.29 & 0.16 & 0.12 & 0.11 & 0.05 & \textbf{0.04} & 0.05 \\
    conv5\_3 & 0.36 & 0.17 & 0.11 & 0.11 & \textbf{0.04} & \textbf{0.04} & \textbf{0.04} \\
    fc6   & \textbf{0.04} & 0.10 & 0.06 & 0.06 & \textbf{0.04} & \textbf{0.04} & \textbf{0.04} \\
    fc7   & 0.04 & 0.11 & 0.05 & 0.06 & \textbf{0.02} & 0.03 & 0.03 \\
    fc8   & 0.23 & 0.22 & 0.11 & 0.12 & 0.08 & \textbf{0.06} & \textbf{0.06}\\
    \hline \hline
  \end{tabular}
  \caption{Layer-wise compression for all 6 coreset techniques (denoted only by their identifiers to save space) on VGGNet-16. The entries represent the fraction of parameters retained post compression.}
  \label{tab:sup2}
\end{table}

\subsection{LeNet-5}
Table~\ref{tab:sup3} provides layer-wise compression performance for LeNet-5 over all 6 techniques.
\begin{table}
  \centering
  \caption{Layer-wise compression for all 6 coreset techniques (denoted only by their identifiers to save space) on LeNet-5. The entries represent the fraction of parameters retained post compression.}
  \begin{tabular}{l||c|c|c|c|c|c|c|c}
    \hline \hline
    \textbf{Layer} & \textbf{Han \textit{et al.}}\cite{han2015learning} & \textbf{Guo \textit{et al.}}\cite{guo2016dynamic} & \textbf{K}    & \textbf{S}    &  \textbf{A}   & \textbf{AP+K} & \textbf{AP+S} & \textbf{AP+A} \\ \hline
    conv1 & 0.66 & 0.14 & 0.06 & 0.03 & 0.03 & 0.02 & \textbf{0.02} & 0.02 \\
    conv2 & 0.12 & 0.03 & 0.04 & 0.03 & 0.03 & 0.02 & \textbf{0.02} & 0.02\\
    fc1 & 0.08 & \textbf{0.01} & 0.04 & 0.03 & 0.03 & 0.02 & \textbf{0.01} & 0.02 \\
    fc2   & 0.19 & 0.04 & 0.02 & \textbf{0.01} & 0.02 & \textbf{0.01} & \textbf{0.01} & \textbf{0.01} \\
    \hline \hline
  \end{tabular}
  \label{tab:sup3}
\end{table}

\section{Visualization}
Additionally, we provide some \textbf{conv1} filter visualizations to showcase the impact of the Coreset compression technique. For the purpose of visualization, we choose Coreset-S, which exhibits the highest compression for several different CNNs. The following figures show \textit{both} the original and the modified \textbf{conv1} filters (upon applying the Coreset-S compression algorithm) for AlexNet, ResNet-18, ResNet-50, and ResNet-101.
\begin{figure}[ht]
  \centering
  \includegraphics[width=\linewidth]{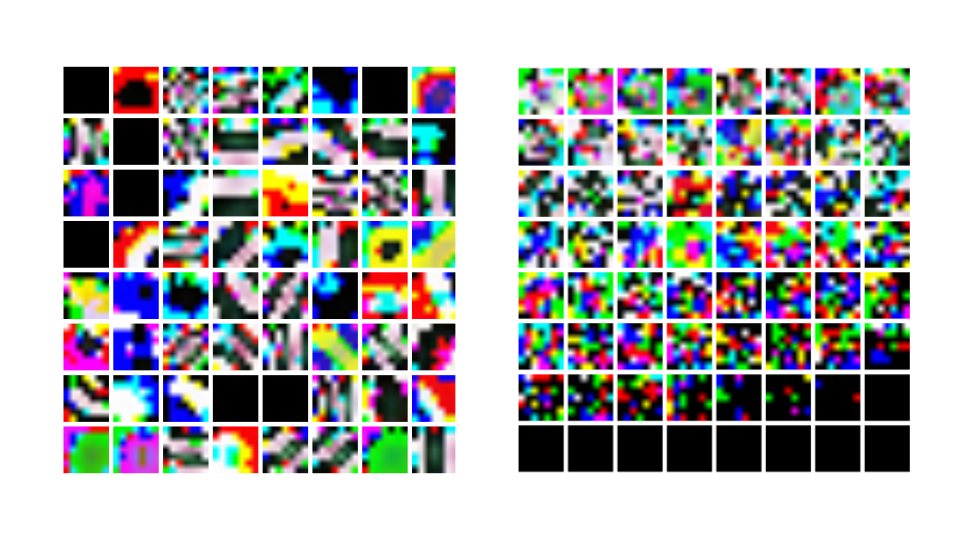}
  \caption{Visualization of \textbf{conv1} filters for ResNet-18. The image on the left represents the original filters learnt through backpropagation, and the image on the right displays the complete Coreset-S representation, ordered from top-left to bottom-right on the basis of their eigenvalues.}
  \label{fig:vis_resnet18}
\end{figure}

\begin{figure}[t]
  \centering
  \includegraphics[width=\linewidth]{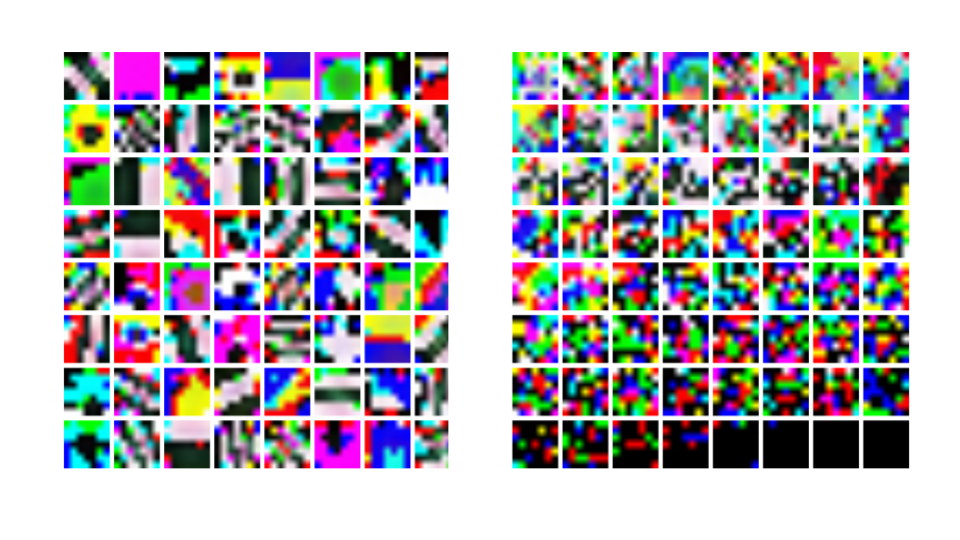}
  \caption{Visualization of \textbf{conv1} filters for ResNet-50. The image on the left represents the original filters learnt through backpropagation, and the image on the right displays the complete Coreset-S representation, ordered from top-left to bottom-right on the basis of their eigenvalues. }
  \label{fig:vis_resnet50}
\end{figure}

\begin{figure}[t]
  \centering
  \includegraphics[width=\linewidth]{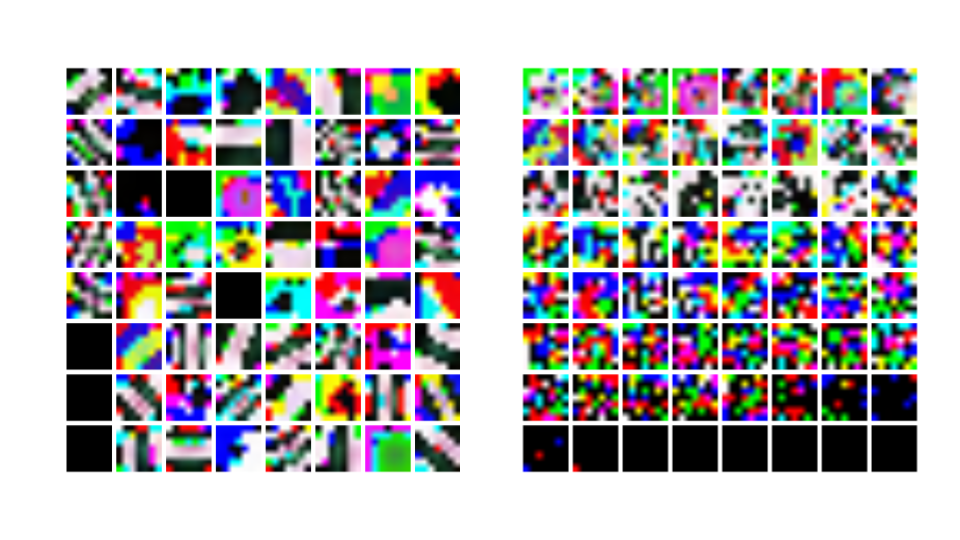}
  \caption{Visualization of the \textbf{conv1} filters for ResNet-101. The image on the left represents the original filters learnt through backpropagation, and the image on the right displays the complete Coreset-S representation, ordered from top-left to bottom-right on the basis of their eigenvalues.}
  \label{fig:vis_resnet101}
\end{figure}

\bibliographystyle{splncs04}
\bibliography{egbib}

%
%
%

\end{document}